\documentclass[journal]{IEEEtran}
\usepackage{subfigure}
\usepackage{graphicx}
\usepackage{xcolor}
\usepackage{amsmath}
\usepackage{mathtools}
\usepackage{amssymb}
\usepackage{verbatim} 
\usepackage{diagbox}
\usepackage{threeparttable} 
\usepackage[normalem]{ulem} 
\useunder{\uline}{\ul}{} 

\usepackage{hyperref} 
\usepackage{booktabs}

\ifCLASSINFOpdf
\else
\fi

\begin{document}
%

\title{Subjective and Objective Analysis of Streamed Gaming Videos}

%
%
%

 \author{Xiangxu Yu, Zhenqiang Ying, Neil Birkbeck, Yilin Wang, Balu Adsumilli and Alan C. Bovik
 \thanks{X. Yu, Z. Ying and A. C. Bovik are with the Department of Electrical and Computer Engineering, University of Texas at Austin, Austin, USA (e-mail: yuxiangxu@utexas.edu; zqying@utexas.edu; bovik@ece.utexas.edu). N. Birkbeck, Y. Wang and B. Adsumilli are with YouTube Media Algorithms Team, Google Inc (e-mail: birkbeck@google.com; yilin@google.com; badsumilli@google.com).}
 }

\markboth{----------------------------------------------}%
{--------------------------------}
%



\maketitle


\begin{abstract}

The rising popularity of online User-Generated-Content (UGC) in the form of streamed and shared videos, has hastened the development of perceptual Video Quality Assessment (VQA) models, which can be used to help optimize their delivery. 
Gaming videos, which are a relatively new type of UGC videos, are created when skilled gamers post videos of their gameplay. 
These kinds of screenshots of UGC gameplay videos have become extremely popular on major streaming platforms like YouTube and Twitch. 
Synthetically-generated gaming content presents challenges to existing VQA algorithms, including those based on natural scene/video statistics models. 
Synthetically generated gaming content presents different statistical behavior than naturalistic videos. 
A number of studies have been directed towards understanding the perceptual characteristics of professionally generated gaming videos arising in gaming video streaming, online gaming, and cloud gaming. 
However, little work has been done on understanding the quality of UGC gaming videos, and how it can be characterized and predicted. 
Towards boosting the progress of gaming video VQA model development, we conducted a comprehensive study of subjective and objective VQA models on UGC gaming videos. 
To do this, we created a novel UGC gaming video resource, called the LIVE-YouTube Gaming video quality (LIVE-YT-Gaming) database, comprised of 600 real UGC gaming videos. 
We conducted a subjective human study on this data, yielding 18,600 human quality ratings recorded by 61 human subjects. 
We also evaluated a number of state-of-the-art (SOTA) VQA models on the new database, including a new one, called GAME-VQP, based on both natural video statistics and CNN-learned features. 
To help support work in this field, we are making the new LIVE-YT-Gaming Database, along with code for GAME-VQP, publicly available through the link: \url{https://live.ece.utexas.edu/research/LIVE-YT-Gaming/index.html}.

\end{abstract}


%
\IEEEpeerreviewmaketitle

\section{Introduction}
\label{introduction_game}

Over the past few years, Internet traffic has continued to grow dramatically. 
According to \cite{ciscociscovisualnetworkingindex}, video streaming now occupies the majority of Internet bandwidth and is the main driver of this growth. 
Streaming service providers like Netflix, Hulu, and Amazon Prime Video generate and stream a substantial portion of this visual traffic in the form of Professionally-Generated-Content (PGC) videos. 
At the same time, online video sharing platforms like YouTube, Vimeo, and Twitch collect and make available vast numbers of User-Generated-Content (UGC) videos, generally uploaded by less skilled videographers. 
Driven by the rapid development of the digital game industry and by consumer desire to see skilled gamers in action, a large number of online videos of gameplay content have been uploaded and made public on the Internet. 
In 2020, YouTube Gaming reached a milestone of 100 billion hours of watch time and 40 million active gaming channels \cite{youtubegamingbiggestyear}. 
The rapid expansion of the gaming market has also given birth to many streaming-related services, such as game live broadcasts, online games, and cloud games. 
At the same time, increasingly powerful computing, communication, and display hardware and software technologies have continuously uplifted the quality of these services. 
In order to provide users with better quality streaming gaming video services and to improve users’ viewing experiences, perceptual Video Quality Assessment (VQA) research has become particularly important. 
Both general-purpose algorithms able to handle diverse distortion scenarios, as well as VQA models specifically designed for gaming videos need to be studied and developed. 

VQA research efforts can generally be divided into two categories: conducting subjective quality assessment experiments, and developing objective quality assessment models and algorithms. 
If it were feasible, subjective quality assessment would best meet the needs of streaming providers and users, but the cost is prohibitive and does not allow for real-time quality based decisions or processing steps, such as perceptually optimized compression. 
Therefore, automatic perception-based objective video quality prediction algorithms are of great interest. 
Objective VQA models can be classified into three categories, based on whether there is a reference video available. 
The first, Full-Reference (FR) models, require the availability of an entire source video as a reference against which to measure visual differences, i.e. video fidelity. 
Reduced-Reference (RR) models only require partial reference information, often only a small amount. 
No-Reference (NR), or ``blind'' models predict the quality of a test video without using any reference information. 
It is generally believed that when predicting the quality of the test video, if there is an available PGC video as a reference, that is, an original, ``pristine'' video of high quality, shot by experts with professional equipment, then FR or RR VQA algorithms are able to achieve superior prediction results, as compared to NR models. 
However, there is often a lack of accessible reference videos, e.g., for UGC videos captured by naive users having limited capture skills or technology. 
These can only be evaluated using NR VQA models. 
Since these kinds of videos are quite pervasive on social media and sharing platforms, the NR VQA problem is of high practical importance. 
Since they also typically encompass much wider ranges of quality and of distortion types which are often simultaneously present and commingled, the NR VQA problem is much more challenging. 

Unlike real-world videos taken by videographers with optical cameras, gaming videos are generally synthetically generated. 
Because of underlying statistical differences in the structure of synthetic and artificial videos \cite{barman2018comparative, barman2018objective}, popular NR VQA models may not achieve adequate quality prediction performance on them. 

Research on the assessment of gaming video quality mostly commenced very recently. 
The authors of \cite{barman2018comparative} studied and analyzed differences between gaming and non-gaming content videos, in terms of compression and user Quality of Experience (QoE). 
Some VQA models designed for gaming videos have been proposed, e.g., \cite{zadtootaghaj2020demi,utke2020ndnetgaming}. 
However, this work has been limited to the study of the characteristics of PGC gaming videos and how they are affected by compression. 
None of these have discussed or analyzed the quality assessment of UGC gaming videos recorded by amateur users. 
We briefly elaborate on the characteristics of PGC and UGC gaming videos in Section \ref{pgc_ugc_gaming_video}. 

Advancements of research on video quality have always relied on freely available databases of distorted videos and subjective judgments of them. 
Over the past two decades, many VQA databases have been created and shared with researchers, which has significantly driven the development of VQA models. 
Databases created in the early stages of VQA research usually contained only a small number of PGC reference videos and a small number of synthetically distorted versions of them. 
In recent years, as UGC VQA research has gained more attention, more advanced and specialized databases have been created, containing hundreds or thousands of unique UGC videos with numerous, complex authentic distortions. 
Recently, databases dedicated to the development of gaming video quality assessment research have also been proposed. 
However, these databases only contain a limited number of PGC gaming videos as references, along with compressed versions of them. 
Therefore, towards advancing progress on understanding streamed UGC gaming video quality, we have created a new UGC gaming video resource, which we call the LIVE-YouTube Gaming video quality (LIVE-YT-Gaming) database. 
This new dataset contains subjective rating scores on a large number of unique gaming videos from a study that was conducted online. 

We summarize the contributions we make as follows: 
\begin{itemize}
  \item Using this new resource, we conducted a novel online human study whereby we collected a large number of subjective quality labels on gaming videos. To demonstrate the utility of the new database, we compared and contrasted the performance of a wide range of state-of-the-art (SOTA) VQA models, including one of our own design that attains top performance. 
  \item We constructed a novel subjective UGC gaming video database, which we call the LIVE-YouTube Gaming video quality database (LIVE-YT-Gaming). The new database contains 600 UGC gaming videos of unique content, from 59 different games. This is the largest video quality database addressing UGC gaming, and it includes the most exemplar games. Unlike the few existing gaming video quality databases, where PGC videos were professionally recorded, the LIVE-YT-Gaming videos were collected from online uploads by casual users (UGC videos) and are generally afflicted by highly diverse, mixed distortions. 
\end{itemize}

The rest of the paper is organized as follows: we discuss related work in Section \ref{related_work}. 
We present the new distorted gaming video database in Section \ref{database}. 
We provide details on the subjective study in Section \ref{study}, and an analysis of the collected subjective data in Section \ref{data_process}. 
We compare the performances of several current VQA models on the new database in Section \ref{performance}, and introduce a new model, called GAME-VQP. 
Finally, we summarize the paper and discuss possible future work in Section \ref{conclusion}. 

\section{Related Work}
\label{related_work}

\subsection{Subjective Video Quality Database}
\label{related_work_database}

Existing VQA databases differ by the types of video contents and distortions that are included, as well by the volume and methods of procuring subjective data on them. 
Two different methods are commonly used to collect subjective ratings on videos in the databases: laboratory studies and online crowdsourced studies. 

\subsubsection{Laboratory VQA Studies}

Most of the early VQA databases collected subjective data in the laboratory. 
A small set of unique reference videos (usually 10 to 20), along with distorted versions of them were presented to a relatively small ($<$100) set of volunteer subjects. 
We will refer to these synthetically-distorted datasets as \textit{legacy} VQA databases. 
Some representative legacy databases are LIVE VQA \cite{seshadrinathan2010study}, LIVE Mobile \cite{moorthy2012video}, CDVL \cite{pinson2013consumer}, MCL-V \cite{lin2015mcl}, MCL-JCV \cite{wang2016mcl}, and VideoSet \cite{wang2017videoset}. 

Laboratory studies have advantages: the human data that is obtained is generally more reliable, and the scientist has significant control over the experimental environment. 
However, it is difficult to recruit a large number of subjects, hence much less data can be collected. 
Because of this, legacy databases are data-poor although they are valuable test beds because of the high quality of their data. 

\subsubsection{Crowdsourced VQA Studies}

In recent years, crowdsourcing as a tool for conducting subjective studies has been used to create a variety of large-scale VQA databases containing UGC videos. 
Crowdsourcing is usually conducted on online platforms like Amazon Mechanical Turk (MTurk) and CrowdFlower, whereby it is possible to collect a large amount of subjective data from participants around the world. 
Crowdsourced databases usually contain hundreds to thousands of UGC videos selected and sampled from online video sources, on which large amounts of subjective data can be collected by crowdsourcing. 
Some representative crowdsourced video quality databases are CVD2014 \cite{nuutinen2016cvd2014}, LIVE-In-Capture \cite{ghadiyaram2017capture}, KoNViD-1k \cite{hosu2017konstanz}, YFCC100M \cite{thomee2016yfcc100m}, LIVE-VQC \cite{sinno2018large}, and YouTube-UGC \cite{wang2019youtube}. 

Crowdsourcing makes it possible to collect a large volume subjective data from thousands of workers. 
However, collected data is less reliable than quality scores collected in the laboratory, and disingenuous subjects must be addressed. 
Further, the researcher must ensure that the subjects have adequate displays and network bandwidths to properly participate. 
Since a very large number of paid workers are required, online studies can be very expensive. 

\subsubsection{Gaming Video Quality Assessment Databases}
\label{gaming_database_compare}

As far as we are aware, there are four video quality databases that have been designed for gaming video quality research purposes: GamingVideoSET \cite{barman2018gamingvideoset}, KUGVD \cite{barman2019no}, CGVDS \cite{zadtootaghaj2020quality}, and TGV \cite{wen2021subjective}. 

The GamingVideoSET database contains 24 gaming contents recorded from 12 different games, along with 576 compressed versions of these videos using H.264 compression. 
However, only 90 of these have associated subjective data. 
Three different resolutions are included: 480p, 720p and 1080p, all with frame rates of 30 frames/second (fps). 
The KUGVD database is similar to GamingVideoSET, but only has 6 gaming contents and 144 compressed versions of them, 90 of which have subjective data available. 
Both GamingVideoSET and KUGVD databases are therefore limited by having small amounts of subjective data, which hinders model development. 
All of the reference videos were recorded under professional conditions without visible distortion, unlike UGC videos.
Moreover, the distorted videos were all compressed using the same H.264 standard, making quality prediction less difficult. 
The CGVDS database was specifically developed for analyzing and modeling the quality of gaming videos compressed using hardware accelerated implementations of H.264/MPEG-AVC. 
It has 360 distorted videos with subjective quality ratings, compressed from 15 PGC reference videos, at framerates of 20, 30, and 60 fps. 
Finally, the TGV database is a mobile gaming video database containing 1293 gaming video sequences compressed from 150 source videos. 
Unlike the aforementioned databases, which only contain computer or console games, the videos in the TGV database were all recorded from 17 different mobile games. 
All four of the above-described gaming video databases contain only PGC videos that have been impaired by a single distortion type (compression). 
Prior to our effort here, there has been no database designed for conducting VQA research on real UGC gaming videos created by casual users. 

A comparison of the four existing gaming video quality databases is given in Table \ref{database_comparison}. 
There is one more recently published database called GamingHDRVideoSET \cite{barman2021user}, which includes HDR gaming videos, however it does not include any subjective data, so we excluded it from our evaluations.

\begin{table*}
  \resizebox{\textwidth}{!}{%
  \begin{threeparttable}
\caption{Evaluation of Four Existing Gaming Video Quality Databases: GamingVideoSET, KUGVD, CGVDS, and TGV}
\label{database_comparison}
\begin{tabular}{cccccccccccccccc}
\toprule
Database           & Year & Content No  & Video No  & Game No  & Subjective Data       & Public & Resolution              & FPS        & Duration & Format      & Distortion Type       & Subject No  & Rating No   & Data        & Study Type   \\ \midrule
GamingVideoSET     & 2018 & 24          & 600       & 12       & 90                    & Yes    & 480p, 720p, 1080p       & 30         & 30 sec   & mp4, yuv    & H.264 compression     & 25          & 25          & MOS         & In-lab study \\ 
KUGVD              & 2019 & 6           & 150       & 6        & 90                    & Yes    & 480p, 720p, 1080p       & 30         & 30 sec   & mp4, yuv    & H.264 compression     & 17          & 17          & MOS         & In-lab study \\ 
CGVDS              & 2020 & 15          & 255       & 15       & 360 + anchor stimuli  & Yes    & 480p, 720p, 1080p       & 20, 30, 60 & 30 sec   & mp4, yuv    & H.264 compression     & over 100    & Unavailable & MOS         & In-lab study \\ 
TGV                & 2021 & 150         & 1293      & 17       & Unavailable           & No     & 480p, 720p, 1080p       & 30         & 5  sec   & Unavailable & H264, H265, Tencent codec & 19      & Unavailable & Unavailable & In-lab study \\ \bottomrule
\end{tabular}
\begin{tablenotes}
\small
 \item Content No: Total number of unique contents. \qquad Video No: Total number of videos. \qquad Game No: Total number of source games. \qquad Subjective Data: Total number of videos with subjective ratings available. \qquad \\ FPS: Framerate per second. \qquad Subject No: Total number of participating subjects. \qquad Rating No: Average number of ratings per video.    
\end{tablenotes}
\end{threeparttable}
}
\end{table*}

\subsection{Objective Quality Assessment Model}

The following is a brief introduction to the development of modern NR VQA models, including both general-purpose models and gaming video quality models. 

\subsubsection{General NR VQA Models}

While the earliest NR VQA models were developed to address specific types of distortion, the focus of VQA research shifted toward the development of more general-purpose NR algorithms, which extract various ``quality-aware'' hand-crafted features, which are fed to simple regressors, such as SVRs, to learn mappings to human subjective quality labels. 

The most successful features derive from models of natural natural scene statistics (NSS) \cite{ruderman1994statistics} and natural video statistics (NVS) \cite{soundararajan2012video}, under which high quality optical images, when subjected to perceptually-relevant bandpass processes obey certain statistical regularities that are predictably altered by distortions. 
Noteworthy examples include NIQE \cite{mittal2013making}, BRISQUE \cite{mittal2012no}, V-BLIINDS \cite{saad2014blind}, HIGRADE \cite{kundu2017no}, GM-LOG \cite{xue2014blind}, DESIQUE \cite{zhang2013no}, and FRIQUEE \cite{ghadiyaram2017perceptual}. 
More recent models that employ efficiently optimized NSS/NVS features, and/or combined with deep features, include VIDEVAL \cite{tu2021ugc}, which leverages a hand-optimized selection of statistical features, and RAPIQUE \cite{tu2021rapique}, which combines a large number of easily-computed statistics with semantic deep features in an extremely efficient manner, yielding both excellent quality prediction accuracy and rapid computation. 

Unlike NSS-based models, data-driven methods like CORNIA \cite{ye2012unsupervised} begin by constructing a codebook via unsupervised learning techniques, instead of using a fixed set of features, followed by temporal hysteresis pooling. 
A recent published NR VQA model called TLVQM \cite{korhonen2019two} makes use of a two-level feature extraction mechanism to achieve efficient computation of a set of distortion-relevant features that measure motion, specific distortion artifacts, and aesthetics. 

Given the emergence of large-scale video quality databases in recent years, the availability of sufficient quantities of training samples has made it possible to train high-performance deep learning models. 
VSFA \cite{li2019quality} makes use of a pre-trained Convolutional Neural Network (CNN) as a deep feature extractor, then integrates the frame-wise features using a gated recurrent unit and a temporal pooling layer. 
It is one of several SOTA deep learning VQA models that attain superior performance on publicly available UGC video quality databases. 
PaQ-2-PiQ \cite{ying2019patches} is a recent frame-based local-to-global deep learning based VQA architecture, while PVQ \cite{ying2020patch} extends the idea of using local quality predictions to improve global space-time VQA performance. 
Other deep models include V-MEON \cite{liu2018end}, NIMA \cite{talebi2018nima}, PQR \cite{zeng2018blind}, and DLIQA \cite{hou2015blind}, which deliver high performance on legacy and UGC video quality databases. 

\subsubsection{Gaming VQA Models}
\label{gaming_VQA_intro}

Several NR VQA models have been proposed for gaming. 
In \cite{zadtootaghaj2018nr}, the authors proposed a blind model called NR-GVQM, which trains an SVR model to evaluate the quality of gaming content by extracting nine frame-level features, and by using VMAF scores as proxy ground truth labels. 
The Nofu model proposed in \cite{goring2019nofu} is a learning-based VQA model, which applies center cropping to rapidly compute 12 frame-based features, followed by model training and temporal pooling. 
The authors of \cite{barman2019no} proposed two models. 
Their NR-GVSQI model also uses VMAF scores as training targets, while their NR-GVSQE model uses MOS as the training target. 
They both make use of basic distortion features that measure blockiness, blurriness, contrast, exposure etc., as well as quality scores generated by the NSS-based IQA models BIQI, BRISQUE, and NIQE. 
The ITU-T standard G.1072 \cite{gaming2020methodology} describes a gaming video QoE model based on three quality dimensions: spatial video quality, temporal video quality, and input quality (interaction quality). 
Two recently released models based on deep learning, called NDNetGaming \cite{utke2020ndnetgaming} and DEMI \cite{zadtootaghaj2020demi}, were both developed on a DenseNet framework. 
NDNetGaming is based on a CNN structure trained on VMAF scores as proxy ground truth, then fine-tuned using MOS. 
Quality prediction phase is then accomplished via temporal pooling. 
DEMI uses a CNN architecture similar to NDNetGaming, but focuses on only two types of distortions, blurriness and blockiness. 

However, each of these systems were designed and developed on databases containing only a small number of PGC reference videos impaired by a single type of manually applied compression. 
None have been systematically trained and tested on UGC gaming videos. 
One reason for this is that there is no UGC gaming video quality database currently available.

\subsection{Video Game and Gaming Video}
\label{introduction_videogame}

Video games generally refer to interactive games that run on electronic media platforms. 
Popular mainstream games may take the form of computer games, console games, mobile games, handheld games, VR games, cloud games, and so on. 
Many games are available on multiple platforms. 
Games also fall into various genres, including role-playing games, adventure games, action games, first-person shooters, real-time strategy games, fighting games, board games, massive multiplayer online role-playing games, and others. 
At present, most gaming video quality research has focused on streamed gaming, such as interactive cloud games and online games, as well as passive live broadcast and recorded gameplay \cite{barman2018evaluation}. 

\subsection{Recording of PGC and UGC Gaming Video}
\label{pgc_ugc_gaming_video}

Next we describe ways by which gaming videos are created, leading to a discussion of PGC and UGC gaming videos. 

Unlike natural videos captured by optical cameras, gaming \textit{videos}, as opposed to live gameplay, are usually obtained by recording the screen of the devices on which the games are being played. 
A variety of factors affect the output quality of recorded gaming videos. 
The most basic are the graphics configuration settings of the operation system, and the settings built into each game. 
The display quality of games is usually affected by settings like spatial resolution, frame rate, motion blur, control of screen-space ambient occlusions, vertical synchronization, variation of texture/material, grass quality, anisotropic filtering, depth of field, reflection quality, water surface quality, shadow quality/style, type of tessellation, and use of triple buffering and anti-aliasing. 
Professional players enjoy the ultimate gaming experience by optimizing these settings, however, they present formidably complicated choices for most players, so games provide more intuitive and convenient options. 
Players can simply set the game quality to high, medium, or low, letting the game program automatically set the appropriate parameters. 

Of course, whether or not the video quality determined by the game settings can be correctly displayed depends heavily on the hardware configuration. 
A sufficiently powerful GPU can provide complete real-time calculation support to ensure that the game display quality is stable at a high level, but ordinary GPUs may not meet all the quality requirements of all users. 
For example, high motion scenes may present with noticeable frame drops or lagging. 
Some games require very large data calculations, so if the GPU cannot process them quickly enough, delays in the display may occur. 
For those games which require connection to the Internet, poor network conditions may also cause the game screen to delay or even freeze. 

The quality of recorded gaming videos is also affected by the recording software used. 
Professional software can accurately reproduce the original gameplay, but substantial system resources are required, increasing the burden on the gameplay device. 
The software may provide options for faster and easier use, such as automatic resolution selection, frame rate reduction, or further compression during recording, which may degrade the output quality. 

When we refer to PGC gaming videos, we refer to videos captured using professional-grade screen recording software, where the video settings during the gameplay are adjusted to a high level, processed and displayed using professional hardware equipment. 
By contrast, UGC gaming videos refer to videos recorded by ordinary, casual non-professional users on ordinary computers or consoles, using diverse game graphics settings, and various types of recording software. 
Thus, unlike PGC videos, recorded UGC game videos recorded may present with a very wide range of perceptual qualities. 

\section{LIVE-YouTube Gaming Video Quality Database}
\label{database}

\begin{figure}
\centering
\includegraphics[width = 1\columnwidth]{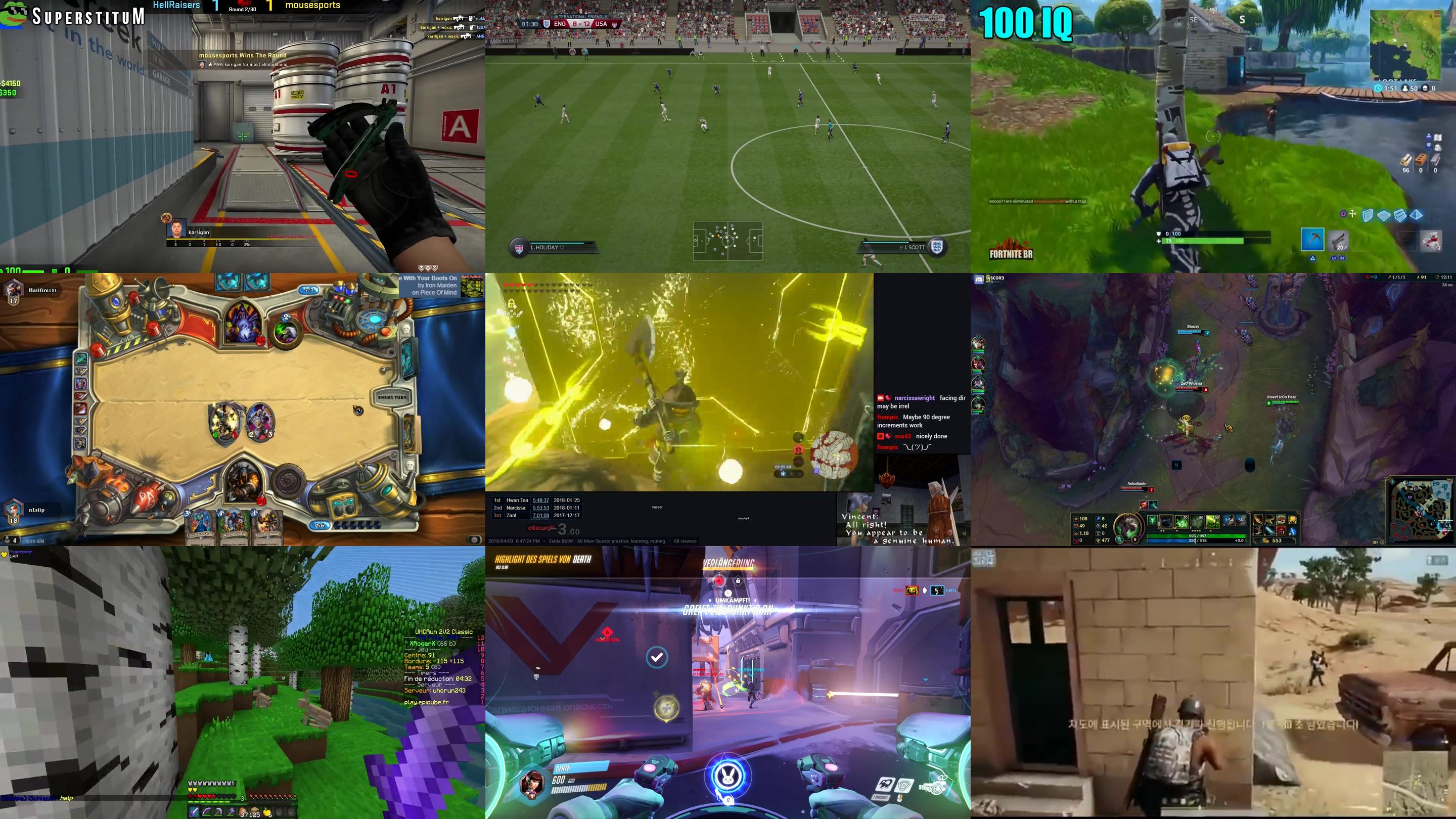}
\caption{Example frames of videos from the new LIVE-YT-Gaming database.}
\label{gamevideo_screenshot}
\end{figure}

We present a detailed description of the new LIVE-YT-Gaming database in this section. 
The new database contains 600 UGC gaming videos harvested from online sources. 
It is the largest public-domain real UGC gaming video quality database having associated subjective scores. 
Our main objective has been to create a resource that will support and boost gaming video quality research. 
By providing this useful and accessible tool, we hope more researchers will be able to engage in research on the perceptual quality of gaming videos. 
Fig. \ref{gamevideo_screenshot} shows a few example videos from the new database. 

Many recent (non-gaming) UGC video quality databases \cite{hosu2017konstanz, ying2019patches, ying2020patch, li2020ugc} were created by harvesting a large number of source videos from one or more large free public video repositories, such as the Internet Archive \cite{internetarchive} or YFCC-100M \cite{thomee2016yfcc100m}. 
These are typically ``winnowed'' to a set of videos that are representative of a category of interest, such as social media videos. 
This is accomplished by a statistical matching process based on low-level video features, such as blur, colorfulness \cite{hasler2003measuring}, contrast, Spatial Information (SI) \cite{winkler2012analysis} and Temporal Information (TI) \cite{itu910subjective}. 
Statistical sampling based on matching is not suitable for harvesting gaming videos, however, because of a dearth of large gaming video databases available for free download. 
While there are numerous and diverse gaming videos that have been uploaded by users onto video websites like YouTube and Twitch, these generally cannot be downloaded because of copyright issues. 
Furthermore, gaming videos are strongly characterized by the type and content of the original games, which impacts the statistical structure of the video signals, such as their bandpass properties \cite{barman2018comparative}. 
Overall, data collection is more difficult and complex than it is for real-world UGC videos. 

\subsection{Video Collection}

We found the Internet Archive (IA), a free digital library, to be a good source of gaming videos. 
Taking into account the popularity of games on YouTube, as well as the wide variety of types of games (as described in Section \ref{introduction_videogame}), we selected 59 games to be included in our database. 
Unlike the completely random downloading of videos used by ourselves and others to create many PGC and UGC databases, we found that the best approach was to search the videos by their game titles, then to ensure the diversity of sources and to reduce bias, videos of each same game title were downloaded randomly, subject to wild resolution and frame rate constraints. 
Four video resolutions were allowed: 360p (360x640), 480p (480x854), 720p (720x1280), and 1080p (1080x1920), and two frame rates were allowed: 30 fps and 60 fps. 
In addition, we used the Windows 10 Xbox game bar \cite{xboxgamebar} to capture the gameplay of some games, at frame rates of 30 fps and 60 fps, and resolutions of 720p and 1080p. 
The video resolutions were selected based on the YouTube video display resolution and aspect ratio standard \cite{youtuberesolution}. 
In the end, we downloaded dozens to hundreds of source videos of each game to use as a data corpus for the next step, video selection. 

\subsection{Video Selection}

After obtaining the source video resources, we randomly extracted a few 10-second clips from each video, thereby obtaining about 3,000 gaming video clips. 
Considering the desired scale of the online human study to be conducted, and the number of subjects available (to be explained), we further randomly selected 600 videos from amongst those clips. 
We then observed the remaining videos, and replaced videos of some popular games that were over-represented, with videos of less popular games to ensure diversity, and we removed videos containing any inappropriate content. 
For a variety of reasons, including avoiding video stalls and limiting the subjects' session durations (Section \ref{stall_resolution_issue}), we cropped the videos to durations in the range of 8-9 sec. 

A summary of the distributions of the video resolutions present in the LIVE-YT-Gaming database is tabulated in Table \ref{vid_resolution}. 

\begin{table}
\caption{Distribution of Video Resolutions in LIVE-YT-Gaming Database}
\label{vid_resolution}
 \centering
\begin{tabular}{ccccc}
\toprule
Resolution & 1080p & 720p & 480p & 360p \\ \midrule
30 fps     & 137   & 187  & 36   & 55   \\ 
60 fps     & 129   & 51   & 0    & 5    \\ \bottomrule
\end{tabular}
\end{table}

\begin{figure}
\centering
\includegraphics[width = 1\columnwidth]{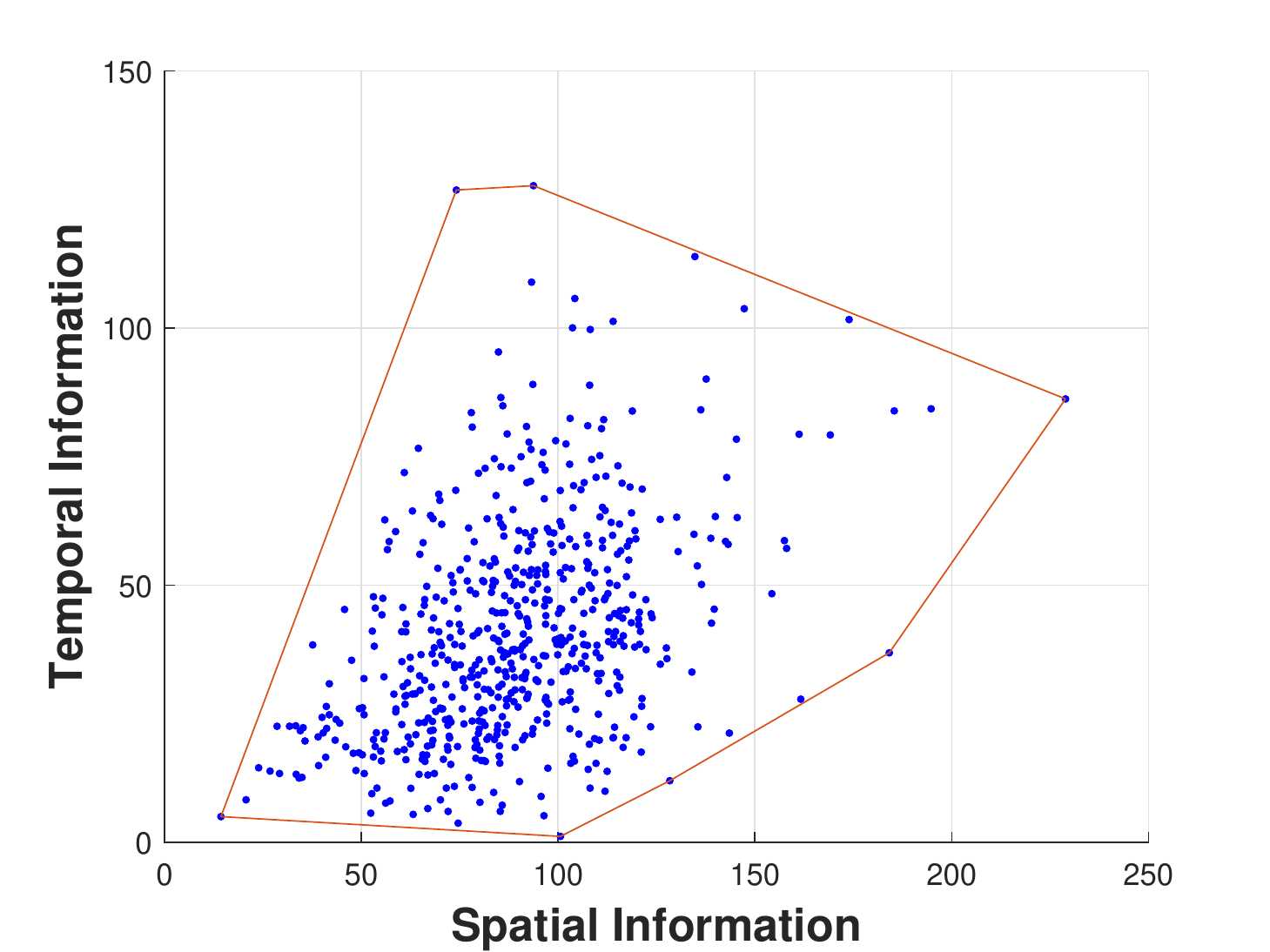}
\caption{Scatter plot of SI against TI on the 600 gaming videos in the LIVE-YouTube Gaming Video Quality Database. }
\label{SI_TI_gaming}
\end{figure}

We also computed SI and TI on all of the remaining videos. 
These quantities roughly measure the spatial and temporal richness and variety of the video contents. 
SI and TI are defined as follows:

\begin{equation}
SI = \textrm{max}_{\mathit{time}}\left \{ std_{space}\left [ \textit{Sobel}( F_{n}(i, j)) \right ] \right \},
\label{SI}
\end{equation}

\begin{equation}
TI = \textrm{max}_{\mathit{time}}\left \{ std_{space}\left [  M_{n}(i, j) \right ] \right \},
\label{TI}
\end{equation}

\noindent
where $F_{n}$ denotes the luminance component of a video frame at instant $n$, $(i, j)$ denotes spatial coordinates, and $M_{n} = F_{n} - F_{n+1}$ is the frame difference operation. 
$Sobel(F_{n})$ denotes Sobel filtering \cite{itu910subjective}. 
Fig. \ref{SI_TI_gaming} shows the distributions of SI and TI for the video contents we selected, indicating that in these aspects, the selected videos contain richer contents than many other UGC video and gaming video databases (\cite{tu2021ugc}, \cite{barman2018objective}, \cite{barman2019no}). 

\section{Subjective Quality Assessment}
\label{study}

\begin{figure}
\centering
\includegraphics[width = 1\columnwidth]{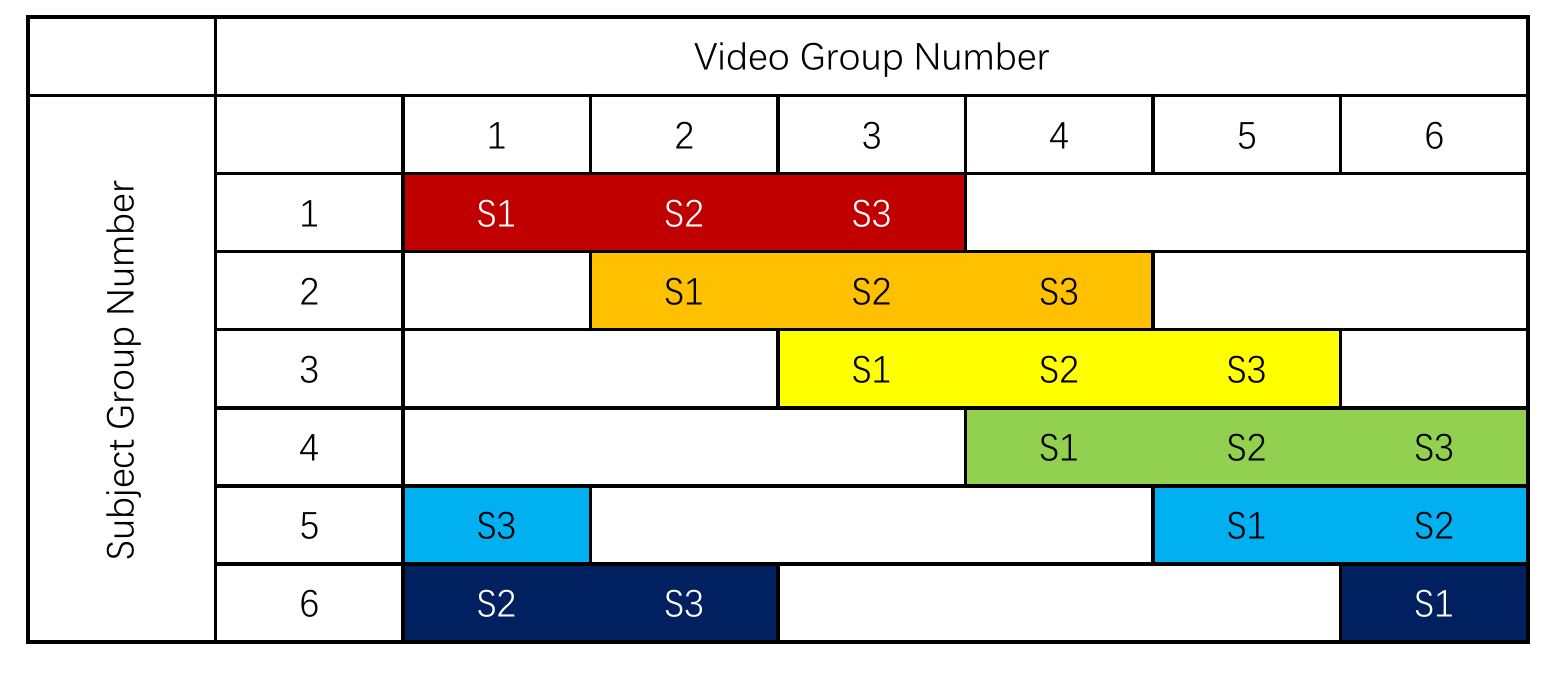}
\caption{Illustration of the round-robin approach used to allocate video groups and subject groups. Grids having the same color indicate video groups watched by subjects in the same group. S1, S2 and S3 are session indices. }
\label{round_robin}
\end{figure}

Next we describe the design and implementation of our online study. 
We stored all of the videos on the Amazon S3 cloud server, providing a safe cloud storage service at high Internet speeds, with sufficient bandwidth to ensure satisfactory video loading speed at the client devices of the study participants. 
We recruited 61 volunteer naive subjects who participated in and completed the entire study. 
All were students at The University of Texas at Austin (UT-Austin) with no background knowledge of VQA research. 
We designed the study in this way, as an online study with fewer, but very reliable subjects, because of Covid-19 regulations. 
Before the study, we randomly divided the subjects into six groups, each containing about 10 subjects. 
We also randomly divided the 600 videos into six groups, each containing 100 videos. 
Each subject watched three groups of videos, or 300 videos. 
In order to avoid any possible bias caused by a same group of videos being watched by a fixed group of subjects, we adopted a round-robin presentation ordering to cross-assign video groups to different subject groups. 
In this way each video would be watched by about 30 subjects from the three different subject groups. 
Fig. \ref{round_robin} illustrates the structure of this round-robin approach. 

\subsection{Study Protocol}

\begin{figure}
\centering
\includegraphics[width = 1\columnwidth]{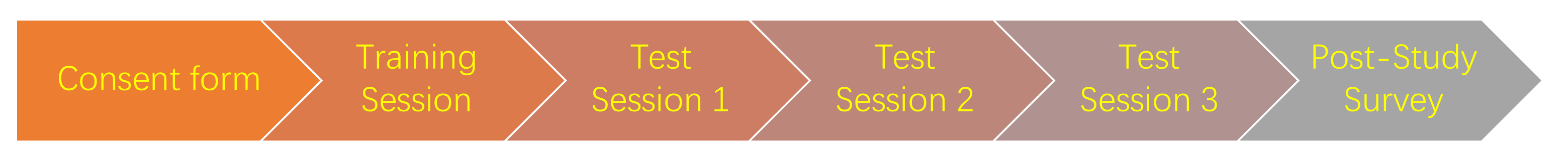}
\caption{Flow chart of the online study. }
\label{study_procedure}
\end{figure}

Fig. \ref{study_procedure} shows the flow chart of the steps of the online study. 
The volunteer subjects first registered by signing a consent form describing the nature of the human study, after which they received an instruction sheet explaining the purpose, procedures, and display device configuration required for the study. 
The subjects were required to use a desktop or laptop computer to complete the experiment, and needed to complete a computer configuration check before the study, to meet the study requirements. 
Each subject received a web link to the training session. 
Based on the data records captured from the training session, we analyzed whether the subjects' hardware configurations met our requirements. 
Following that, the subjects received a link to the three rounds of testing sessions at two-day intervals, three rounds overall. 
After the subjects finished the entire study, they were asked to complete a short questionnaire regarding their opinions of the study.

\subsection{Training Session}

The training session was conducted as follows. 
After opening the link, the subjects first read four webpages of instructions. 
The first webpage introduced the purpose and basic flow of the study. 
There were five sample videos at the bottom of the page, labeled as bad, poor, fair, good, and excellent, respectively, exemplifying the possible quality levels they may encounter in the study. 
The second webpage provided an explanation of how to use the rating bar. 
The third webpage introduced the study schedule and how to submit data. 
The last webpage explained other particulars, such as suggested viewing distance, desired resolution settings, the use of corrective lenses if normally worn, and so on. 
The subjects were required to display each instructional webpage for at 30 seconds before proceeding to the next page. 

\begin{figure}
\centering
\includegraphics[width = 1\columnwidth]{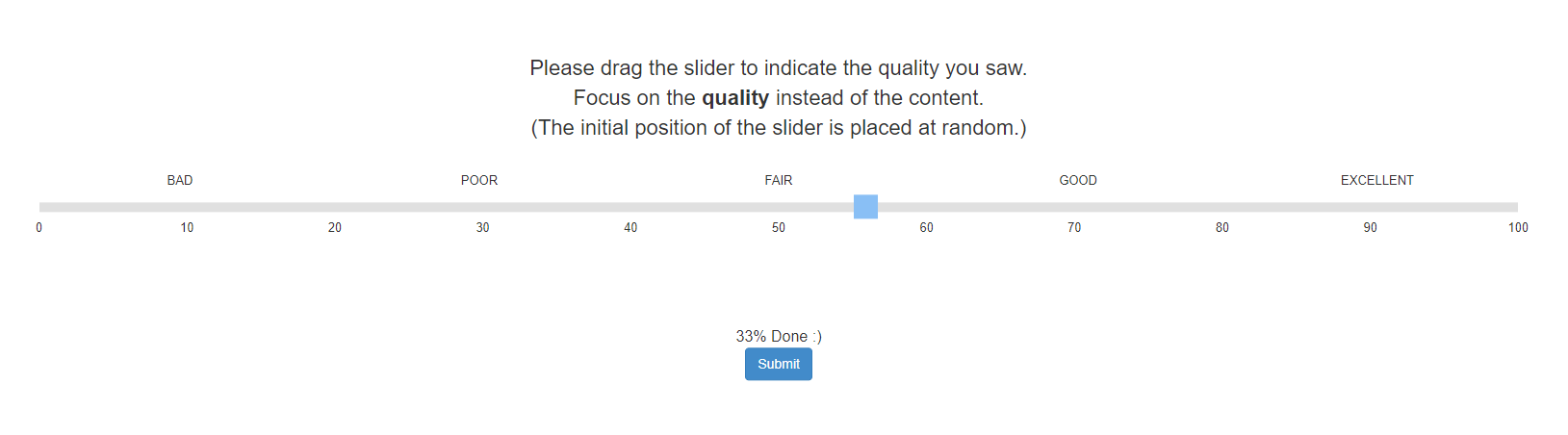}
\caption{Screenshot of the rating bar used in the online human study. }
\label{rating_bar}
\end{figure}

The subjects then entered the experiential training phase and started watching videos. 
After a subject watched a gaming video, a rating bar appeared after it was played. 
It was emphasized that they were to provide ratings of video quality, rather than of content or other aspects. 
On the rating page, a continuous Likert scale \cite{likert1932technique} was displayed, as shown in Fig. \ref{rating_bar}. 
The quality range was labeled from low to high with five guide markers: BAD, POOR, FAIR, GOOD, and EXCELLENT. 
The initial position of the rating cursor was randomized. 
The subject was asked to provide an overall opinion score of video quality by dragging the marker anywhere along the continuous rating bar. 
After the subject clicked the ``Submit" button, the marker's final position was considered to be the rating response. 
Then the next video was presented, and the process repeated until the end of the session. 
The rating scores received from all the subjects were linearly mapped to a numerical quality score in [0, 100]. 

All of the videos presented in each session were displayed in a random order, and each appeared only once. 
The order of presentation was different across subjects. 

\subsection{Test Session}

Each subject participated in a total of three test sessions, each about 30 minutes. 
The subjects were allowed to take a break in the middle of the session, if they desired. 
The sessions were provided to each subject on alternating days to avoid fatigue and memory bias. 
The steps of the test session were similar to the training session, except that there was no time limit on viewing of the instruction page, so subjects could quickly browse and skip the instructions. 

\subsection{Post Questionnaire}

We received feedback from 53 of the participants regarding aspects of the study. 

\subsubsection{Video Duration}
We asked the subjects for their opinions on the video playback duration. 
A summary of the results is given in Table \ref{question_duration}. 
Among the subjects who participated in the questionnaire, 79\% believed that the durations of the observed videos (8-9 seconds) was long enough for them to accurately rate the video quality. 
The results in the Table indicate that the video durations were generally deemed to be satisfactory. 

\begin{table}
\centering
\caption{The Opinion of Study Participants About Video Duration}
\label{question_duration}
\begin{tabular}{cccc}
\toprule
    & Long enough      & Not long enough & Could be shorter \\ \midrule
No. & 42 (79.2\%) & 8 (15.1\%) & 3 (5.7\%)      \\ \bottomrule
\end{tabular}
\end{table}

\subsubsection{Dizziness}

Another issue is that some participants may feel dizzy when watching some gaming videos, especially those that contain fast motion. 
From the survey results in Table \ref{question_dizziness}, about two-thirds of the subjects did not feel any discomfort during the test, while the remaining one-third suffered varying degrees of dizziness. 
This is an important issue that should be considered in other subjective studies of gaming, since it may affect the reliability of the final data. 

\begin{table}
\centering
\caption{Opinions of Study Participants Regarding Video-Induced Dizziness}
\label{question_dizziness}
\resizebox{\columnwidth}{!}{%
\begin{tabular}{cccccc}
\toprule
    & None        & \textless{}30\% & 30\%$\sim$50\% & 50\%$\sim$75\% & \textgreater{}75\% \\ \midrule
No. & 34 (64.2\%) & 16 (30.2\%)     & 2 (3.8\%)      & 1 (1.9\%)      & 0                  \\ \bottomrule
\end{tabular}
}
\end{table}

\subsubsection{Demographics}

\begin{figure}
 \centering
 \subfigure[]{
   \label{watch_video_time}
   \includegraphics[width=0.48\columnwidth]{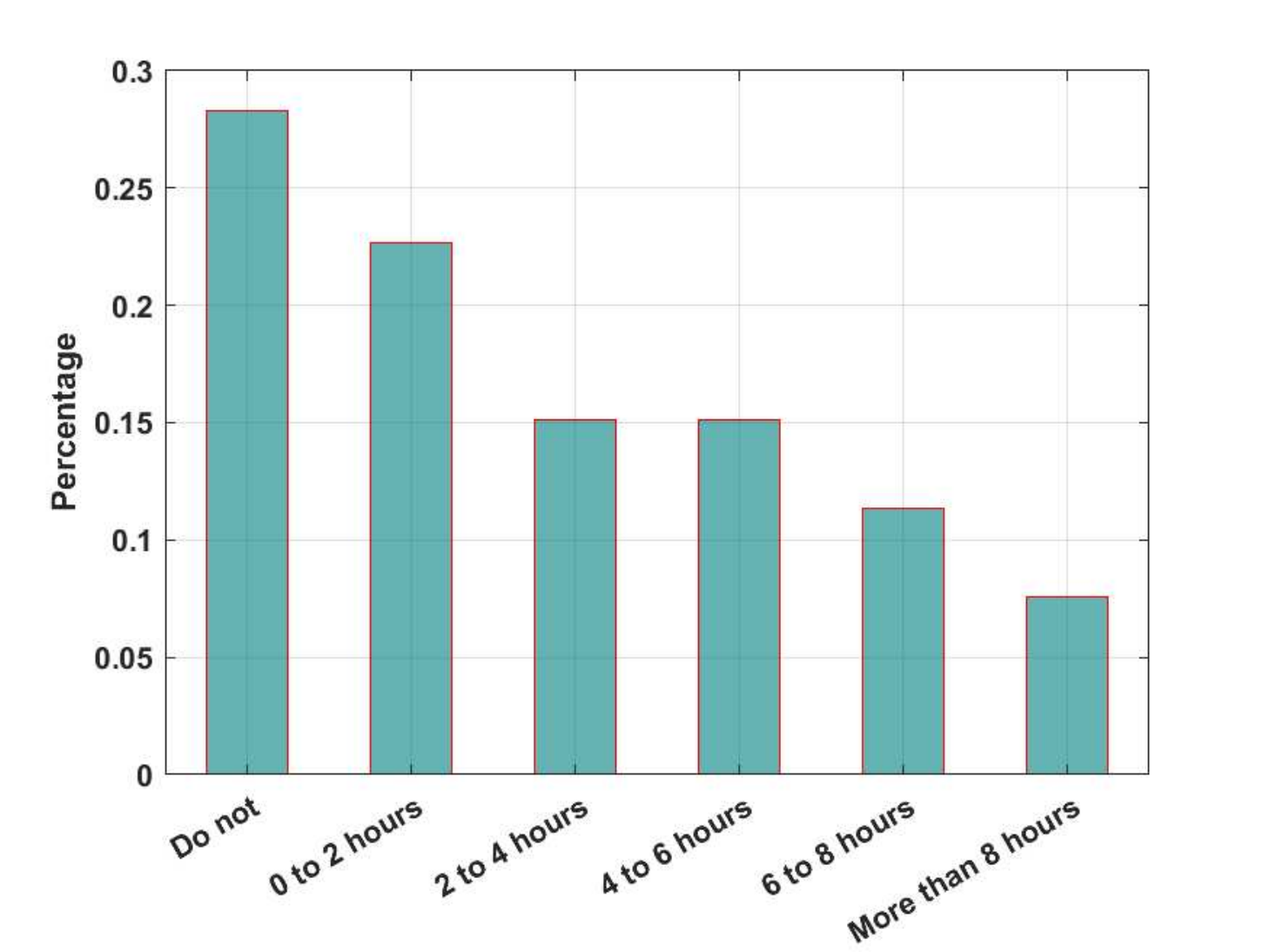}}
 \subfigure[]{
   \label{watch_video_device} 
   \includegraphics[width=0.48\columnwidth]{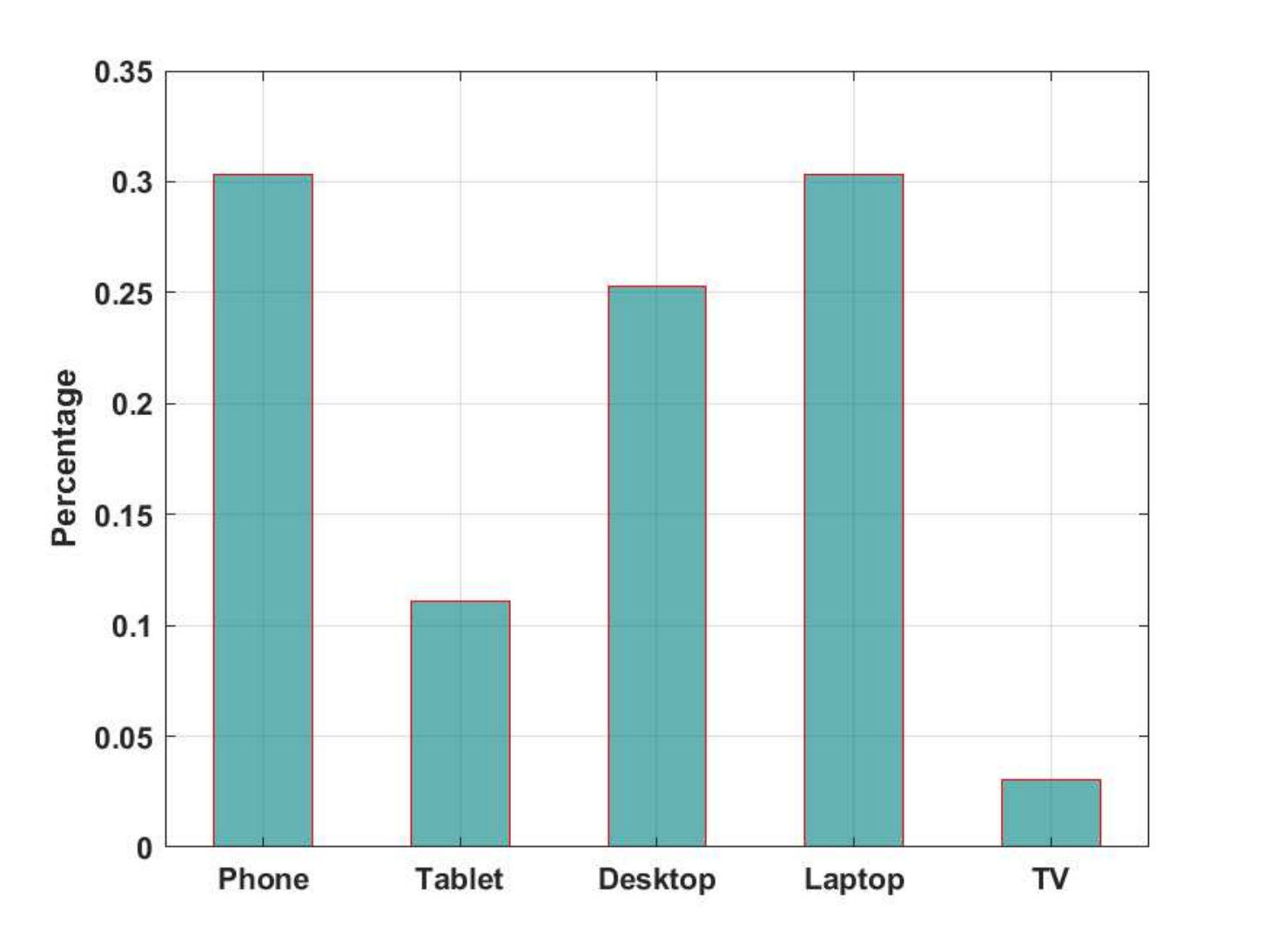}}
 \caption{Demographic details of the participants (a) Typical number of total hours watching gaming videos each week. (b) Device used to watch gaming videos (multiple choice question). }
 \label{watch_video_survey} 
\end{figure}

All of the participants were between 18 to 22 years old. 
Fig. \ref{watch_video_survey} plots the statistics from the answers to two questions: the total time typically spent watching gaming videos each week, and the devices they used to watch gaming videos. 
Approximately 30\% of the participants did not watch gaming videos, while 50\% of them watched at least 2 hours of gaming videos a week. 
Most of the subjects watched gaming videos on computers including laptops and desktops, while 30\% of them watched gaming videos on mobile phones. 
This suggests that it is of interest to conduct additional research on the quality of gaming videos on mobile devices. 

\subsection{Data Recording} 

In addition to the subject’s subjective quality scores, we also recorded information on the subject’s computing equipment, display, network conditions, and real-time playback logs. 
These data helped us guarantee the reliability of the ratings collected. 
Examination of the collected data, along with feedback from the subjects, revealed no issues worth acting upon. 

We also recorded the random initial values of the rating cursor for each displayed video and compared them against the final scores. 
This was done to ensure that the subjects were responsive and moved the cursor. 
We recorded the operating system used by each subject, of the three allowed types: Windows, Linux and macOS, and the model and version of the browser. 

\subsection{Challenges}
\label{stall_resolution_issue}

During our study, we encountered two issues which had to be addressed to ensure the reliability of the collected data. 
The first was video stalls caused by poor Internet conditions. 
If a subject’s Internet was unstable, a video could be delayed at startup or paused during playback. 
This must be avoided, since the subjects might account for any delays, pauses or rebuffering events when giving their quality ratings, leading to inaccurate results. 
The second problem was artifacts arising from automatic rescaling by the client device. 
For example, if a subject were to not set the browser to full screen mode, their device may spatially scale the videos to fit the current window size, which introducing rescaling artifacts which may affect the subjects' video quality scores.  
We took the following steps to deal with these issues. 

\subsubsection{Video Stalls}

To avoid stalls, we applied several protocols. 
First, each video was required to download entirely before playback. 
As mentioned, the videos were of 8-9 sec duration. 
While most of the videos had volumes less than 20 Megabytes (MB), if any exceeded this, we applied very light visually lossless compression to reduce their size. 
In this way, we were able to reduce the burden of the Internet download process. 
Likewise, as each video was playing, download of the next two videos would commence in the background. 
By preloading the videos to be played next, the possibility of video playback problems arising from network instabilities was reduced. 

We also recorded a few relevant parameters of each video playback to determine whether each subject's Internet connection was stable and whether the videos played correctly. 
We recorded the playing time of each video on the subject's device, then compared it with the actual duration of the video, to detect and measure any playback delays. 
We calculated the total (summed) delay times of all videos played in each same session, and found that the accumulated delay time over all sessions of all subjects was less than 0.5 sec, indicating that all of the videos played smoothly. 
We attributed this to the fact that all subjects participated in the same local geographic area (Austin, Texas), avoiding problems encountered in large, international online studies. 

\subsubsection{Rescaling Effects}

On most devices, videos are automatically rescaled to fit the screen size, which can  introduce rescaling artifacts that may significantly alter the perceived video quality, resulting in inaccurate quality ratings. 
To avoid this difficulty, we recorded the resolution setting of each subjects' device as they opened the study webpage. 
We asked all subjects to set their system resolution to 1080x1920, and to display the webpage in full screen mode throughout the study, so the videos would be played at their original resolutions. 
Before each session, we recorded the display resolution of the subject's system and the actual resolution of their browser in full screen mode. 
We also checked the zoom settings of the system and browser and required the subjects to set it to default (no zoom). 
The subjects had to pass these simple criteria before beginning each session. 

\subsection{Comparison of Our Study with Laboratory and Crowdsourced Studies} 

There are interesting similarities and differences between our online study and conventional laboratory and crowdsourced studies. 

\subsubsection{Volume of Collected Data}

The amount of data obtained in the study depends on the number of available videos and of participating subjects. 
Laboratory studies usually accommodate only dozens to hundreds of both subjects and original video contents. 
By contrast, crowdsourced studies often recruit thousands of workers who may collectively rate thousands of videos. 
While our online study was not as large as crowdsourced studies, it was larger than most laboratory VQA studies in regards to both video volume and subject subscription. 

\subsubsection{Study Equipment}

The experimental equipment used in laboratory studies is provided by the researchers, and is often of professional/scientific grade. 
Because of this, laboratory studies can address more extreme scenarios, such as very high resolutions or frame rates, and high dynamic display ranges. 
Crowdsourced studies rely on the subjects' own equipment, which generally implies only moderate playback capabilities, which limits the research objectives. 
In large crowdsourced studies, many of the participants may be quite resource-light. 
These are also constrained by the tools available in the crowdsourcing platform being used. 
For example, MTurk does not allow full-screen video playback. 
Therefore, 1080p videos are often automatically downscaled to adapt to the platform page size, introducing rescaling distortions. 

By comparison, the subject pool in our experiment was composed of motivated university students who already are required to possess adequate personal computing resources and who generally have access to gigabit Internet in the Austin area.

\subsubsection{Reliability}

The reliability of the study mainly depends on control over the experimental environment and on the veracity and focus of the subjects. 
Laboratory studies are conducted in a professional scientific setting, and the recruited subjects generally derive from a known, reliable pool. 
They also personally interact with and are accompanied by the researchers. 
The online study that we conducted was similar, since although it was carried out remotely, a similar reliable subject pool was subscribed, who were in remote communication with the research team in regards to the study instructions and to receive assistance to help them set the environment variables. 
This was felt to be an optimal approach to address Covid restrictions at the time. 
A crowdsourced study is limited by the test platform, and by the often questionable reliability of the remote, unknown subjects, with whom communication is inconvenient. 
Moreover, substantial subject validation and rejection protocols must be desired to address the large number of inadequately equipped, distracted, or even frankly dishonest subjects \cite{ghadiyaram2016massive, ying2019patches, ying2020patch}. 

\section{Subjective Data Processing}
\label{data_process}

We describe the processing of the collected Mean Opinion Score (MOS) next. 
The raw MOS data was first converted into z-scores. 
Let $s_{ijk}$ denote the score provided by the $i$-th subject on the $j$-th video in session $k = \{1, 2, 3\}$. 
Since each video was only rated by half of the subjects, let $\delta(i,j)$ be the indicator function

\begin{equation}
	\delta(i,j) =
	\begin{cases}
		1 & \text{if subject $i$ rated video $j$} \\
		0 & \text{otherwise}. 
	\end{cases}
	\label{indicate_fun}
\end{equation}

The z-scores were then computed as follows: 

\begin{equation}
\begin{aligned}
&z_{ijk} = \frac{s_{ijk} - \bar{s}_{ik}}{\sigma_{ik}},
\end{aligned}
\end{equation}

where

\begin{equation}
\begin{aligned}
&\bar{s}_{ik} = \frac{1}{N_{ik}}\sum_{j=1} ^{N_{ik}} s_{ijk}\\
\end{aligned}
\end{equation}

and

\begin{equation}
\begin{aligned}
&\sigma_{ik} = \sqrt{\frac{1}{N_{ik}-1} \sum_{j=1} ^{N_{ik}} (s_{ijk} - \bar{s}_{ik})^2}, \\
\end{aligned}
\end{equation}

\noindent
where $N_{ik}$ is the number of videos seen by subject $i$ in session $k$. 
The z-scores from all subjects over all sessions were computed to form the matrix $\{z_{ij}\}$, where $z_{ij}$ is the z-score assigned by the $i$-th subject to the $j$-th video, $j \in \{ 1,2 \ldots 600 \}$. 
The entries of $\{z_{ij}\}$ are empty when $\delta(i,j)=0$. 
Subject rejection was conducted to remove outliers, following the recommended procedure described in ITU-R BT 500.13 \cite{series2012methodology}, resulting in 5 of the 61 subjects being rejected. 
The z-scores $z_{ij}$ of the remaining 56 subjects were then linearly rescaled to $[0, 100]$. 
Finally, the MOS of each video was calculated by averaging the rescaled z-scores: 
\begin{equation}
	MOS_j = \frac{1}{N_j} \sum_{i=1} ^N z_{ij}'\delta(i,j),
\end{equation}
where $z_{ij}'$ are the rescaled z-scores, $N_j = \sum_{i=1} ^N \delta(i,j)$, and $N = 600$. 
The MOS values all fell in the range $[4.52, 95.95]$. 

\subsection{Subject-Consistency Test}

To assess the reliability of the collected subjective ratings, we performed inter and intra subject consistency analysis in the following two ways. 

\paragraph{\textit{Inter-Subject Consistency}} 

We randomly divided the subjective ratings obtained on each video into two disjoint equal groups, calculated the MOS of each video, one for each group, and computed the SROCC values between the two randomly divided groups. 
We conducted 100 such random splits, obtaining a median SROCC of \textbf{0.9400}, indicating a high degree of internal consistency. 

\paragraph{\textit{Intra-Subject Consistency}}
The intra-subject reliability test provides a way to measure the degree of consistency of individual subjects \cite{hossfeld2013best}. 
We thus measured the SROCC between the individual opinion scores and the MOS. 
A median SROCC of \textbf{0.7804} was obtained over all of the subjects. 

Both the inter and intra consistency experiments illustrate the high degree of reliability and consistency of the collected subjective ratings. 

\subsection{Analysis}

\begin{figure}
\centering
\includegraphics[width = 1\columnwidth]{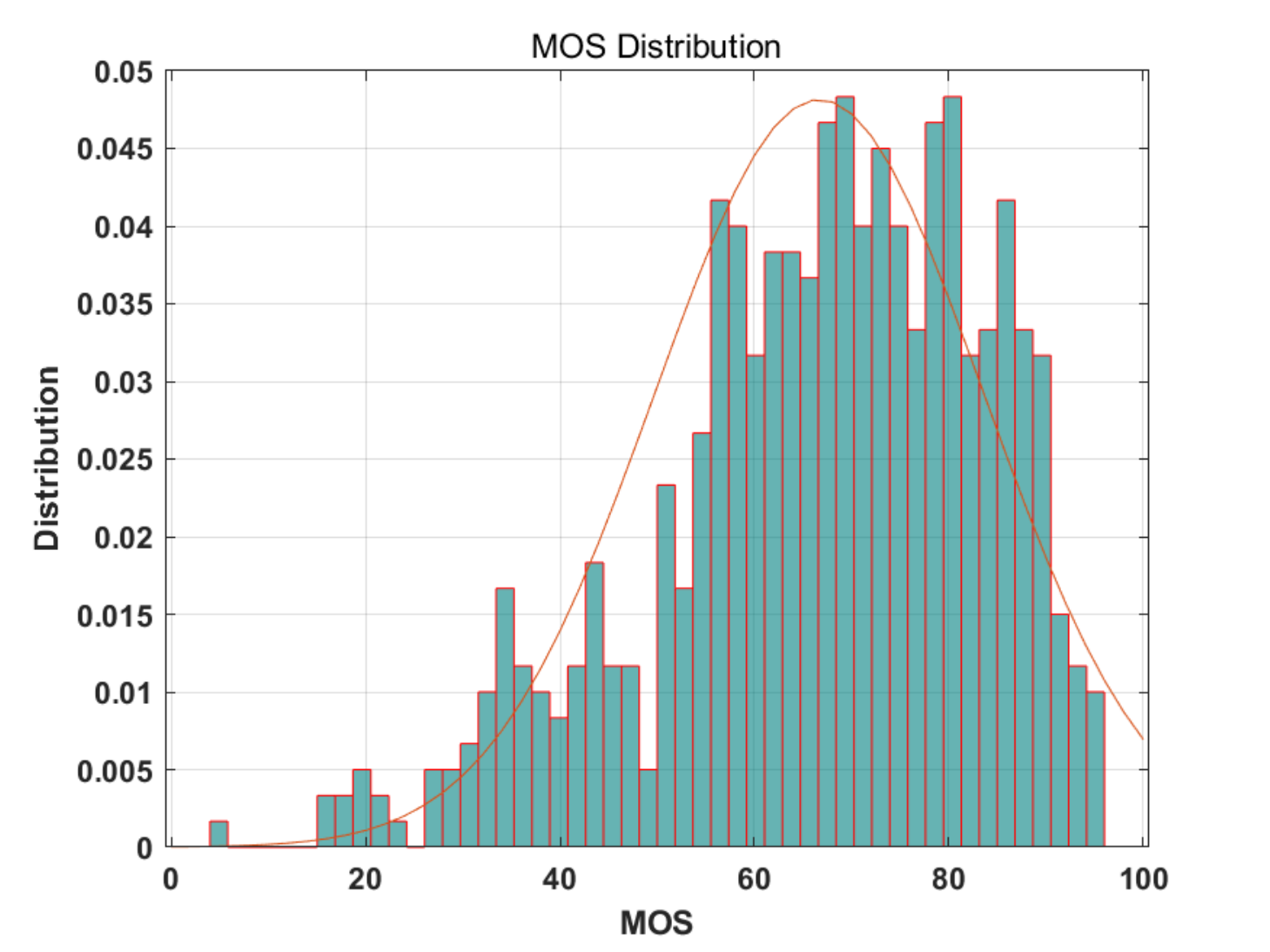}
\caption{MOS distribution across the entire LIVE-YouTube Gaming Video Quality Database. }
\label{MOS_hist}
\end{figure}

\begin{figure}
\centering
\includegraphics[width = 1\columnwidth]{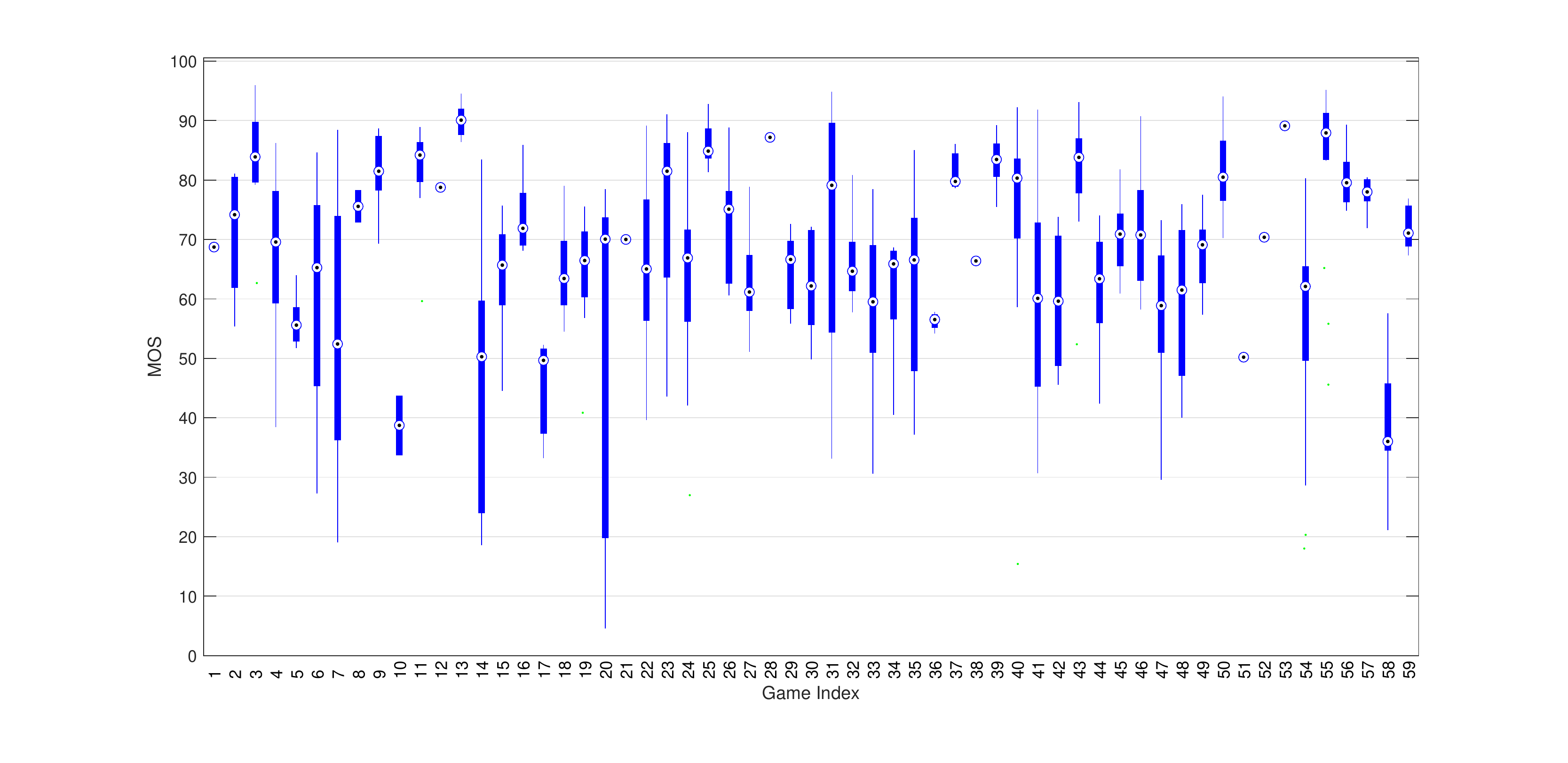}
\caption{Boxplot of the MOS distribution of videos for each game in the LIVE-YouTube Gaming Video Quality Database. The x-axis indexes the different games. }
\label{game_mos_boxplot}
\end{figure}

Table \ref{LIVE-gaming_info} lists the particulars of the LIVE-YT-Gaming database. 
The overall MOS histogram of the database is plotted in Fig. \ref{MOS_hist}, showing a right-skewed distribution. 
Fig. \ref{game_mos_boxplot} shows the boxplots of the MOS of each game, demonstrating the diverse content and quality distribution of the new database. 

\begin{table*}
\caption{Details of the LIVE-YT-Gaming Database}
\label{LIVE-gaming_info}
  \resizebox{\textwidth}{!}{%
\begin{tabular}{cccccccccccccccc}
\toprule
Database       & Year & Content No & Video No & Game No & Subjective Data & Public  & Resolution              & FPS        & Duration & Format   & Distortion Type & Subject No. & Rating No & Data & Study Type   \\ \midrule
LIVE-YT-Gaming    & 2021 & 600        & 600      & 59      & 600             & Yes     & 360p, 480p, 720p, 1080p & 30, 60     & 8-9 sec  & mp4      & UGC distortions & 61          & 30        & MOS  & Online study \\ \bottomrule
\end{tabular}
}
\end{table*}

\section{Performance and Analysis}
\label{performance}

To demonstrate the usefulness of the new LIVE-YT-Gaming database, we evaluated the quality prediction performance of a variety of leading public-domain NR VQA algorithms on it. 
We selected four popular feature-based NR VQA models: BRISQUE, TLVQM, VIDEVAL, and RAPIQUE, all based on training a Support Vector Regression (SVR) \cite{scholkopf2000new}, a `completely blind,' training-free model, NIQE, and a deep learning based model, called VSFA. 
We used the LIBSVM package \cite{chang2011libsvm} to implement the SVR for all algorithms that we retrained on the new databases, with a Radial Basis Function (RBF) kernel. 
We implemented VSFA using the code released by the authors. 
We also tested the performance of two pre-trained networks VGG-16 \cite{simonyan2014very} and Resnet-50 \cite{he2016deep}, by applying the network to the videos and using the output of the fully connected layer to train an SVR model for. 
Since our main goal is to evaluate the quality prediction performance of different algorithms for gaming videos, we also include one gaming video quality model, NDNetGaming, the code of which is publicly available. 
We evaluated the performance between predicted quality scores and MOS using three criteria: Spearman’s rank order correlation coefficient (SROCC), Pearson’s (linear) correlation coefficient (LCC) and the Root Mean Squared Error (RMSE). 
SROCC measures the ranked correlation of two sample distributions without refitting. 
Before computing the LCC and RMSE measures, the predicted quality scores were passed through a logistic non-linearity as described in \cite{VQEG2000}. 
Larger values of both SROCC and LCC imply better performance, while larger values of RMSE indicate worse performance. 

We randomly divided the database into non-overlapping 80\% training and 20\% test sets. 
We repeated the above process over 100 random splits, and report the median performances over all iterations. 
At each iteration, the number of samples in the training set and test set were 480 and 120, respectively. 
For all models, we made use of the publicly available source code from the authors, and used their default settings.

\subsection{A New Blind Gaming VQA Model}

We recently created a blind VQA model designed for the quality prediction of UGC gaming videos, which overcomes the limitations of existing VQA models when evaluating both synthetic and authentic distortions. 
We will refer to this model as the Game Video Quality Predictor, or GAME-VQP. 
Our proposed model utilizes a novel fusion of NSS features and deep learning features to produce reliable quality prediction performance. 
In the design of GAME-VQP, a bag of spatial and spatio-temporal NSS features are extracted over several color spaces, supported by the assumption that NSS features from different spaces capture distinctive aspects of perceived quality. 
A widely used pre-trained CNN model, Resnet-50, was used to extract deep learning features. 
The extracted NSS features and CNN features were each used to train an independent SVR model, then the final prediction score was obtained as the average prediction score of the two models. 
We also include results that compare the performance of GAME-VQP with other leading VQA models. 

\subsection{Evaluation Results}

\begin{table*}
\caption{Performance Comparison of Various No-Reference VQA Models on The LIVE-YouTube Gaming Video Quality Database Using Non-Overlapping 80\% Training And 20\% Test Sets. The Numbers Denote Median Values Over $100$ Iterations of Randomly Chosen Non-Overlapping 80\% Training And 20\% Test Sets (Subjective MOS vs Predicted MOS). The Boldfaces Indicate A Top Performing Model. The Italics Indicate Deep Learning VQA Models. The Underline Indicates The Prior VQA Model Designed for Gaming Videos. }
\label{performance_model}
\resizebox{\textwidth}{!}{%
\begin{tabular}{ccccccccccc}
\toprule
      & NIQE   & BRISQUE & TLVQM  & VIDEVAL & RAPIQUE &  \textit{VSFA} & \textit{VGG-16} & \textit{Resnet-50} & {\ul NDNetGaming}        & \textbf{GAME-VQP} \\ \midrule
SROSS & 0.2801 & 0.6037  & 0.7484 & 0.8071  & 0.8028  & 0.7762         & 0.5768          & 0.7290             & 0.4640                   & \textbf{0.8451}   \\ 
LCC   & 0.3037 & 0.6383  & 0.7564 & 0.8118  & 0.8248  & 0.8014         & 0.6429          & 0.7677             & 0.4682                   & \textbf{0.8649}   \\ 
RMSE  & 16.208 & 13.268  & 11.134 & 10.093  & 9.661   & 10.396         & 13.240          & 11.083             & 15.108                   &  \textbf{8.878}   \\ \bottomrule
\end{tabular}
}
\end{table*}

\begin{table}
\caption{Results of One-Sided Wilcoxon Rank Sum Test Performed Between SROCC Values of The VQA Algorithms Compared In Table \ref{performance_model}. A Value Of "1" Indicates That The Row Algorithm Was Statistically Superior to The Column Algorithm; " $-$ 1" Indicates That the Row Was Worse Than the Column; A Value Of "0" Indicates That the Two Algorithms Were Statistically Indistinguishable. The Boldfaces Indicate The Top Performing Model. The Italics Indicate Deep Learning VQA Models. The Underline Indicates A Prior VQA Model Designed for Gaming Videos. }
\label{performance_statistc_srocc}
\resizebox{\columnwidth}{!}{%
\begin{tabular}{cccccccccccccc}
\toprule
                   & NIQE       & BRISQUE    & TLVQM      & VIDEVAL    & RAPIQUE    & \textit{VSFA} & \textit{VGG-16} & \textit{Resnet-50} & {\ul NDNetGaming}  & \textbf{GAME-VQP} \\ \midrule 
NIQE               & 0          & -1         & -1         & -1         & -1         & -1            & -1              & -1                 & -1                 & \textbf{-1}       \\ 
BRISQUE            & 1          & 0          & -1         & -1         & -1         & -1            & 1               & -1                 & 1                  & \textbf{-1}       \\ 
TLVQM              & 1          & 1          & 0          & -1         & -1         & -1            & 1               & 1                  & 1                  & \textbf{-1}       \\ 
VIDEVAL            & 1          & 1          & 1          & 0          & 0          & 1             & 1               & 1                  & 1                  & \textbf{-1}       \\ 
RAPIQUE            & 1          & 1          & 1          & 0          & 0          & 1             & 1               & 1                  & 1                  & \textbf{-1}       \\ 
\textit{VSFA}      & 1          & 1          & 1          & -1         & -1         & 0             & 1               & 1                  & 1                  & \textbf{-1}       \\ 
\textit{VGG-16}    & 1          & -1         & -1         & -1         & -1         & -1            & 0               & -1                 & 1                  & \textbf{-1}       \\ 
\textit{Resnet-50} & 1          & 1          & -1         & -1         & -1         & -1            & 1               & 0                  & 1                  & \textbf{-1}       \\ 
{\ul NDNetGaming}  & 1          & -1         & -1         & -1         & -1         & -1            & -1              & -1                 & 0                  & \textbf{-1}       \\ 
\textbf{GAME-VQP}  & \textbf{1} & \textbf{1} & \textbf{1} & \textbf{1} & \textbf{1} & \textbf{1}    & \textbf{1}      & \textbf{1}         & \textbf{1}         & \textbf{0}        \\ \bottomrule
\end{tabular}
}
\end{table}

\begin{table}
\caption{Results of One-Sided Wilcoxon Rank Sum Test Performed Between LCC Values of The VQA Algorithms Compared In Table \ref{performance_model}. A Value Of "1" Indicates That The Row Algorithm Was Statistically Superior to The Column Algorithm; " $-$ 1" Indicates That the Row Was Worse Than the Column; A Value Of "0" Indicates That the Two Algorithms Were Statistically Indistinguishable. The Boldfaces Indicate The Top Performing Model. The Italics Indicate Deep Learning VQA Models. The Underline Indicates A Prior VQA Model Designed for Gaming Videos. }
\label{performance_statistc_lcc}
\resizebox{\columnwidth}{!}{%
\begin{tabular}{cccccccccccccc}
\toprule
                   & NIQE       & BRISQUE    & TLVQM      & VIDEVAL    & RAPIQUE    & \textit{VSFA}  & \textit{VGG-16} & \textit{Resnet-50} & {\ul NDNetGaming}  & \textbf{GAME-VQP} \\ \midrule
NIQE               & 0          & -1         & -1         & -1         & -1         & -1             & -1              & -1                 & -1                 & \textbf{-1}       \\ 
BRISQUE            & 1          & 0          & -1         & -1         & -1         & -1             & 1               & -1                 & 1                  & \textbf{-1}       \\ 
TLVQM              & 1          & 1          & 0          & -1         & -1         & -1             & 1               & 1                  & 1                  & \textbf{-1}       \\ 
VIDEVAL            & 1          & 1          & 1          & 0          & -1         & 1              & 1               & 1                  & 1                  & \textbf{-1}       \\ 
RAPIQUE            & 1          & 1          & 1          & 1          & 0          & 1              & 1               & 1                  & 1                  & \textbf{-1}       \\ 
\textit{VSFA}      & 1          & 1          & 1          & -1         & -1         & 0              & 1               & 1                  & 1                  & \textbf{-1}       \\ 
\textit{VGG-16}    & 1          & -1         & -1         & -1         & -1         & -1             & 0               & -1                 & 1                  & \textbf{-1}       \\ 
\textit{Resnet-50} & 1          & 1          & -1         & -1         & -1         & -1             & 1               & 0                  & 1                  & \textbf{-1}       \\ 
{\ul NDNetGaming}  & 1          & -1         & -1         & -1         & -1         & -1             & -1              & -1                 & 0                  & \textbf{-1}       \\ 
\textbf{GAME-VQP}  & \textbf{1} & \textbf{1} & \textbf{1} & \textbf{1} & \textbf{1} & \textbf{1}     & \textbf{1}      & \textbf{1}         & \textbf{1}         & \textbf{0}        \\ \bottomrule
\end{tabular}
}
\end{table}

The performances of all models are shown in Table \ref{performance_model}. 
To determine whether there exists significant differences between the performances of the compared models, we conducted a statistical significance test. 
We used the distributions of the obtained SROCC and LCC values computed over the 100 random train-test iterations. 
The non-parametric Wilcoxon Rank Sum Test \cite{wilcoxon1945individual}, which compares the rank of two lists of samples, was used to conduct hypothesis testing. 
The null hypothesis was that the median for the row model was equal to the median of the column model at the 95\% significance level. 
The alternate hypothesis was that the median of the row was different from the median of the column. 
A value of `1' in the table means that the row algorithm was statistically superior to the column algorithm, while a value of `-1' means the counter result. 
A value of `0' indicates that the row and column algorithms were statistically indistinguishable (or equivalent). 
The statistical significance results are tabulated in Tables \ref{performance_statistc_srocc} and \ref{performance_statistc_lcc}. 
As may be observed, our GAME-VQP performed significantly better than the other algorithms. 
VIDEVAL and RAPIQUE performed well, but still fell short despite their excellent performance on UGC videos. 

The completely blind NR VQA model (NIQE) did not perform well in the new database. 
This is understandable because the pristine model used by NIQE was created using natural pristine images, while gaming videos are synthetically generated and have very different statistical distributions. 
One could imagine a NIQE created on pristine gaming videos. 
Although BRISQUE extracts features similar to NIQE, training it on the UGC gaming videos produced much better results. 
The relative results of NIQE and BRISQUE suggest that the statistical structures of the gaming videos are different from those of natural videos, they nevertheless possess regularities that may be learned using NSS features. 
TLVQM captures motion characteristics from the videos and delivered better performance than BRISQUE, indicating the importance of accounting for motion in gaming videos. 
The performance of the deep learning VSFA model was close to that of RAPIQUE and VIDEVAL, and better than the pre-trained Resnet-50 and VGG-16 models. 
These results show that deep models are able to capture the characteristics of synthetic videos, suggesting the potential of deep models for gaming VQA. 
VIDEVAL and RAPIQUE both performed well, showing the effectiveness of combining NSS features with CNN features. 
This also shows that not all NSS and CNN features successfully contribute to UGC gaming video quality prediction, since the performance of GAME-VQP exceeded those models.  
By selecting fewer features that deliver high performance, our gaming VQA model produced results that were better than all of the other models. 

\begin{figure}
 \centering
 \subfigure[]{
   \label{performance_boxplot:SROCC}
   \includegraphics[width=0.9\columnwidth]{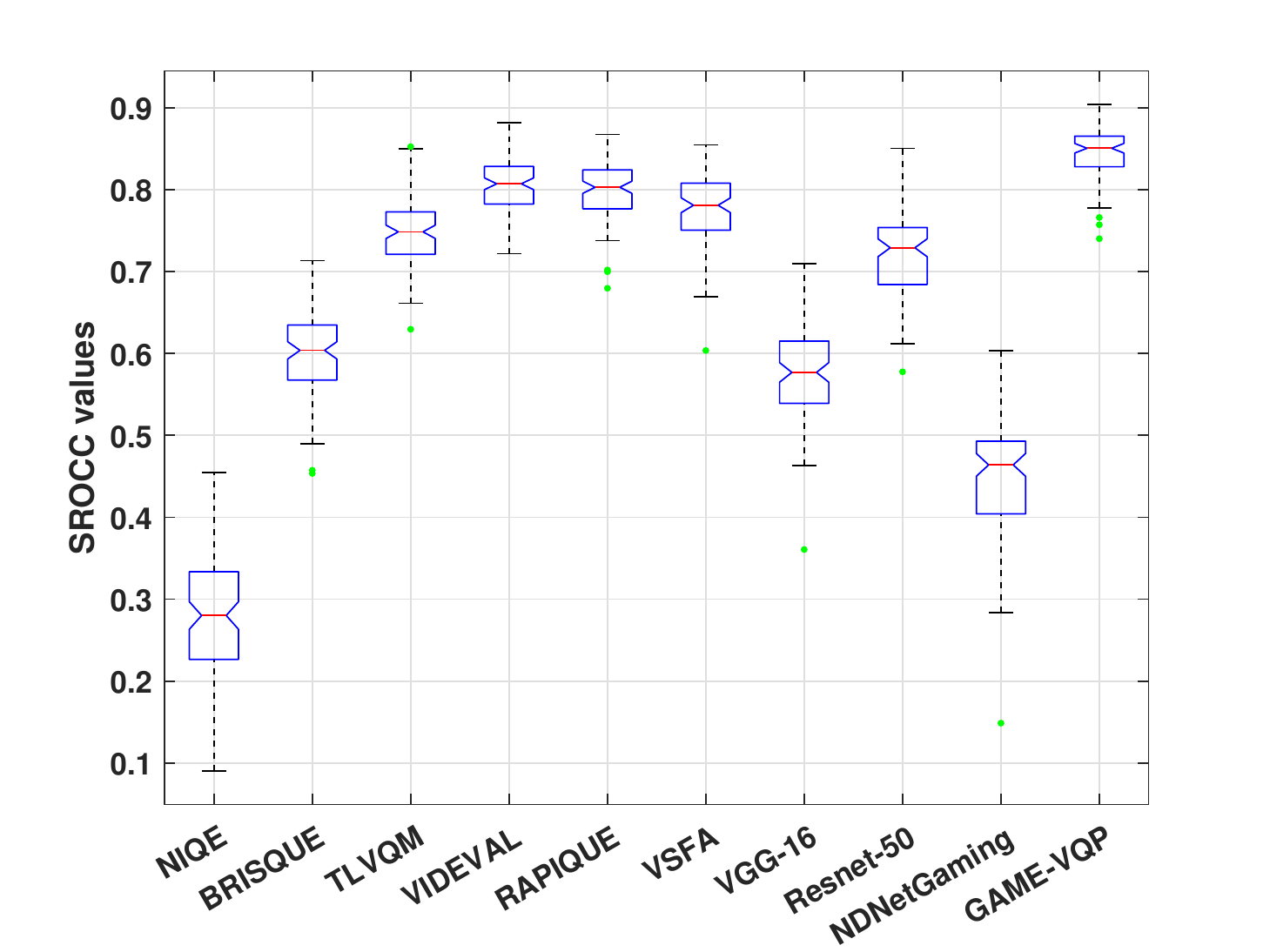}}
 \subfigure[]{
   \label{performance_boxplot:LCC} 
   \includegraphics[width=0.9\columnwidth]{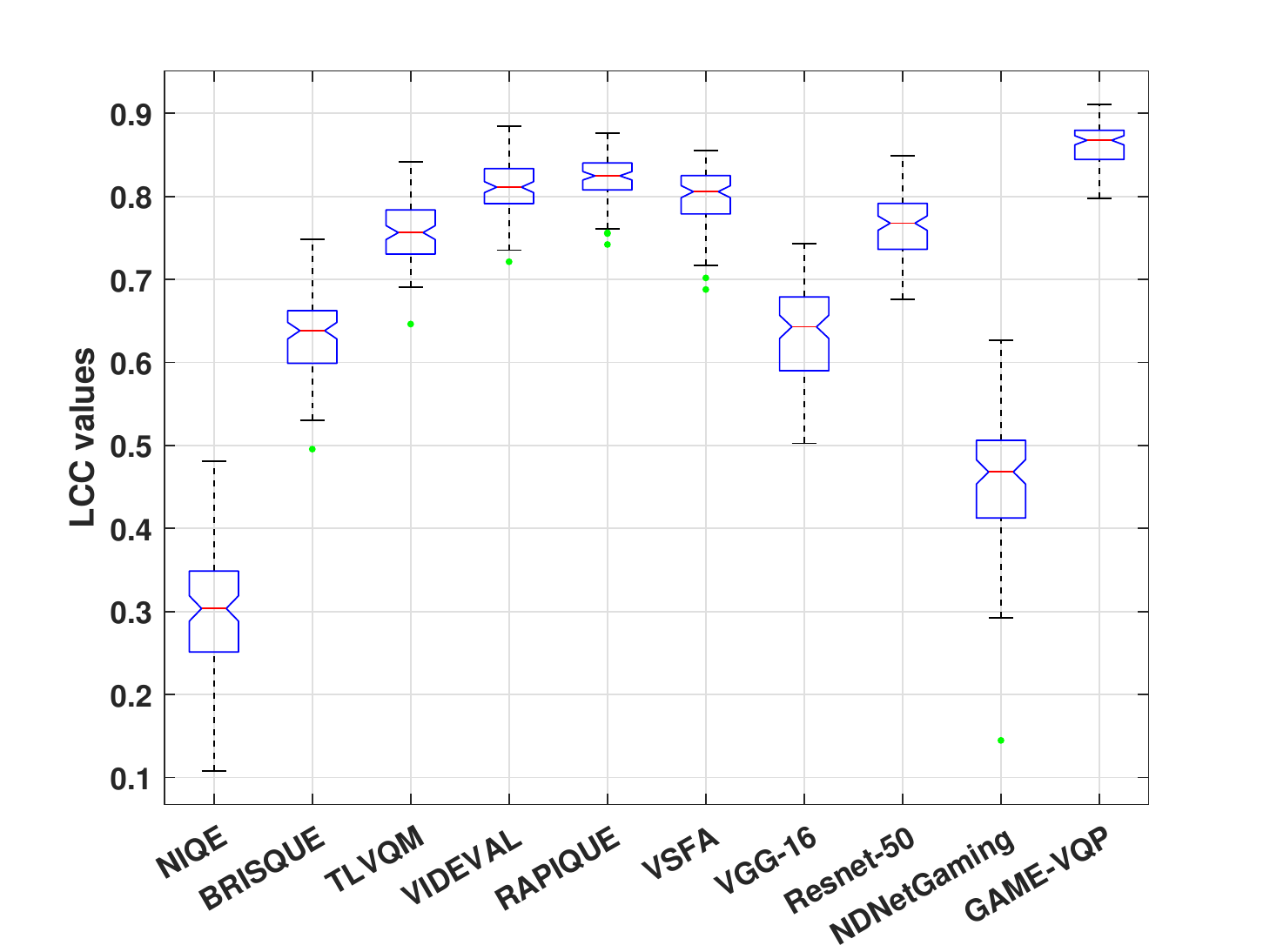}}
 \caption{Box plots of the SROCC and LCC distributions of the algorithms compared in Table \ref{performance_model} over 100 randomized trials on the LIVE-YouTube Gaming Video Quality Database. The central red mark represents the median, while the bottom and top edges of the box indicate the 25th and 75th percentiles, respectively. The whiskers extend to the most extreme data points not considered outliers, while the outliers are individually plotted using the `.' symbol. }
 \label{performance_boxplot} 
\end{figure}

Fig. \ref{performance_boxplot} shows box plots of the SROCC and LCC correlations obtained over 100 iterations for each of the algorithms compared in Table \ref{performance_model}. 
A lower standard deviation with a higher median SROCC or LCC values indicates better and more robust performance. 
Our proposed algorithm clearly exceeded the performance of all the other algorithms, both in terms of stability and performance results. 

\subsection{Scatter Plot}

\begin{figure}
 \centering
 \subfigure[]{
   \label{performance_scatter:NIQE}
   \includegraphics[width=0.48\columnwidth]{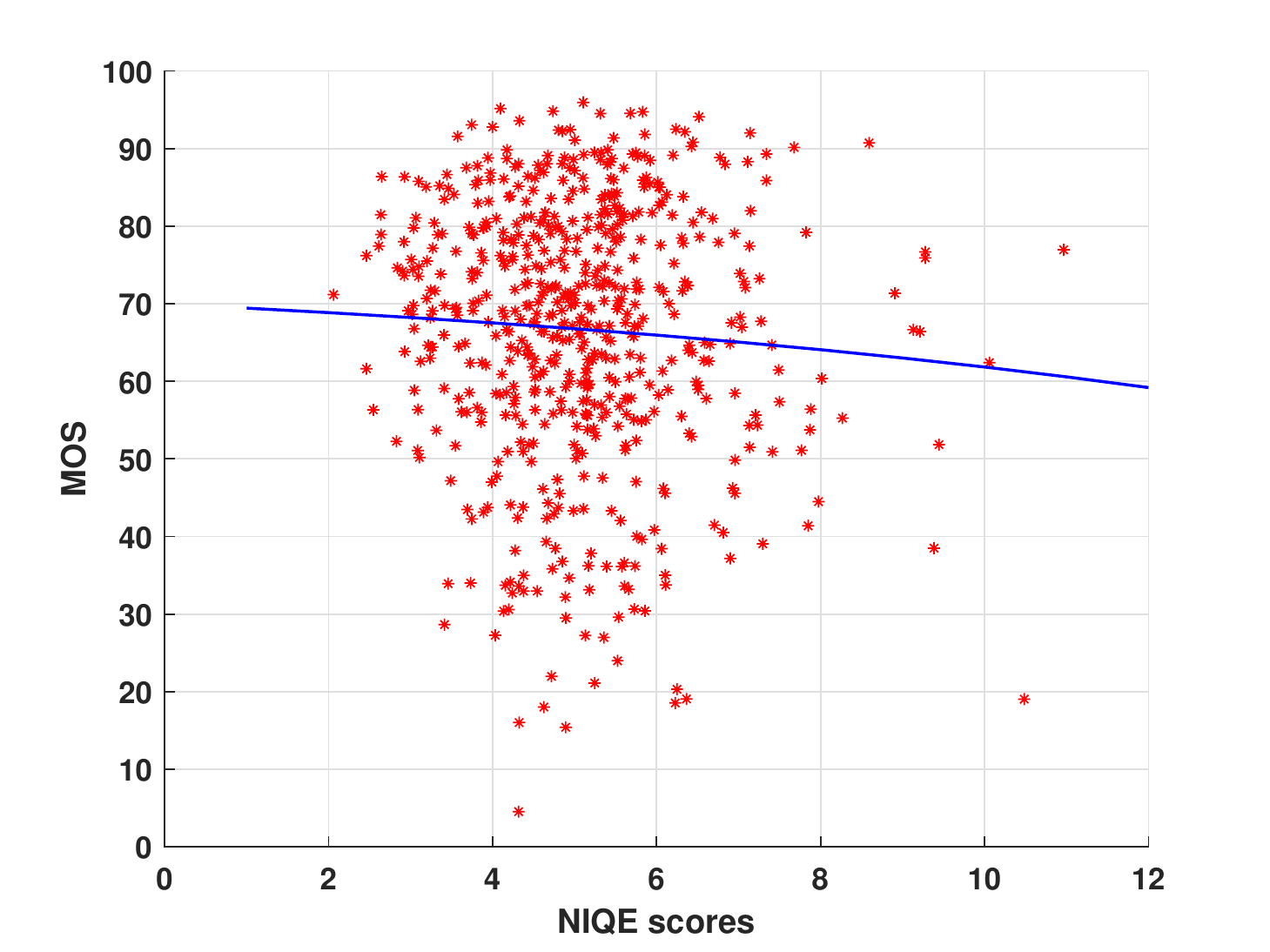}}
 \subfigure[]{
   \label{performance_scatter:TLVQM} 
   \includegraphics[width=0.48\columnwidth]{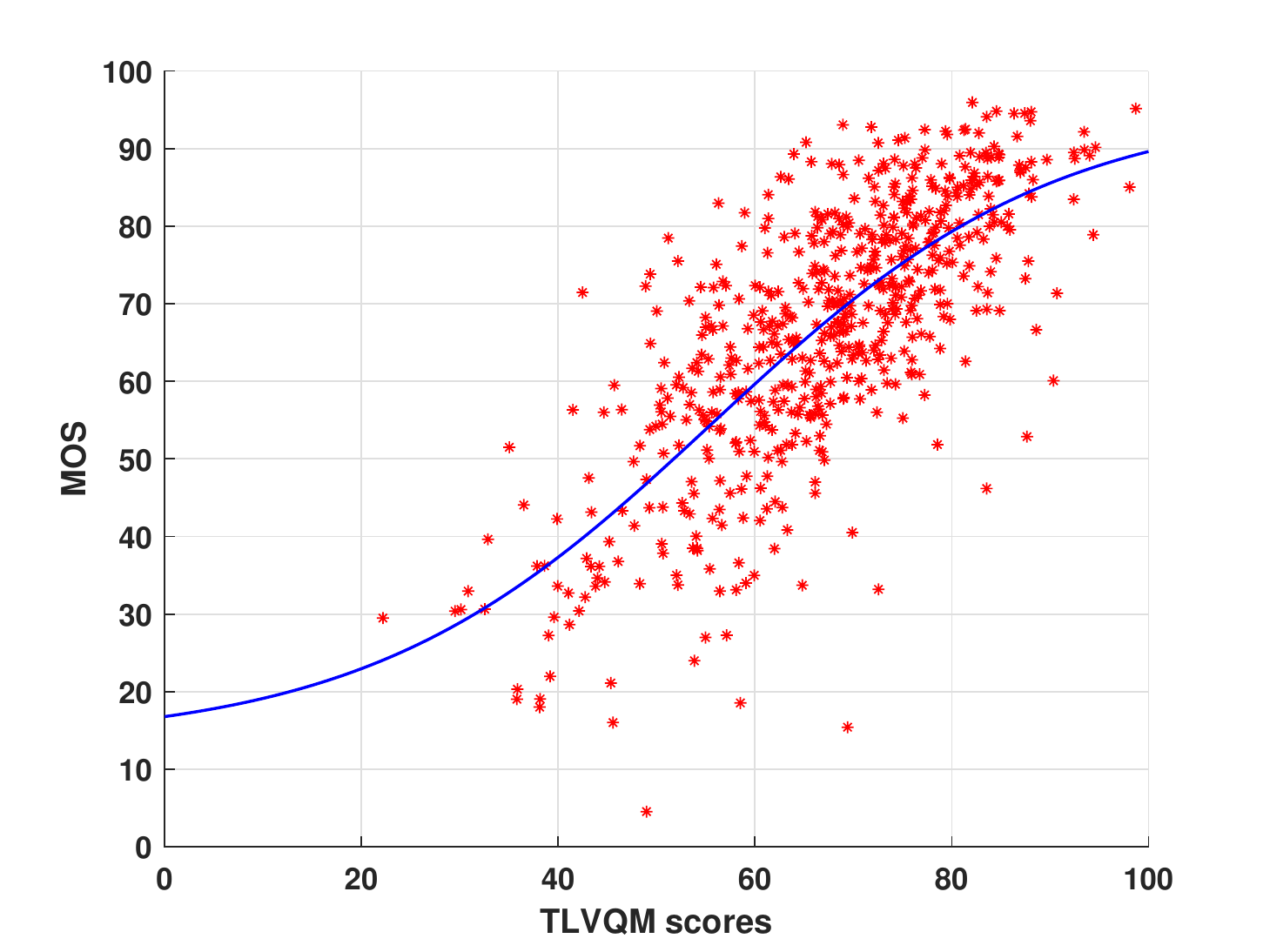}}
 \subfigure[]{
   \label{performance_scatter:RAPIQUE} 
   \includegraphics[width=0.48\columnwidth]{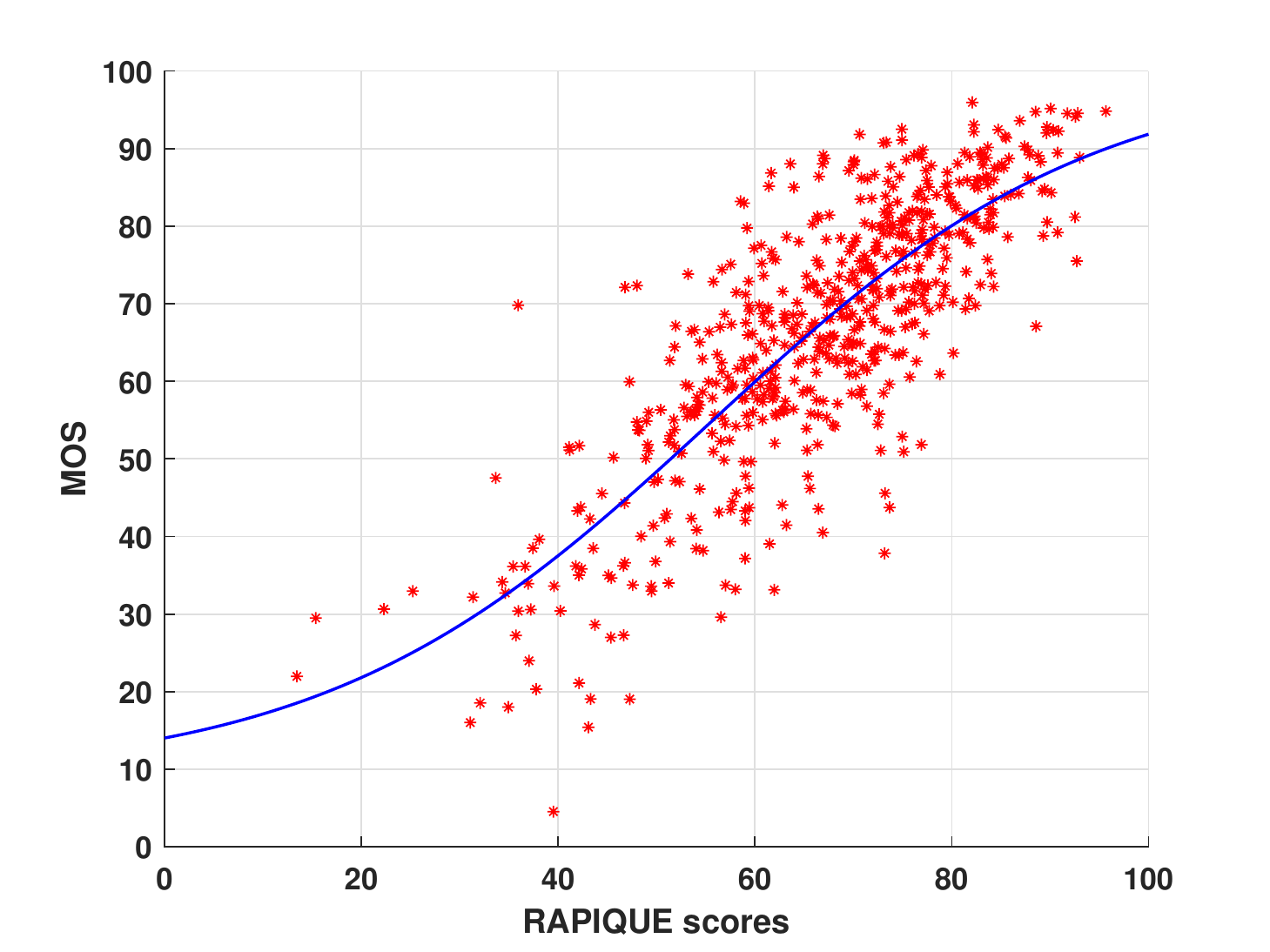}}
 \subfigure[]{
   \label{performance_scatter:Resnet50} 
   \includegraphics[width=0.48\columnwidth]{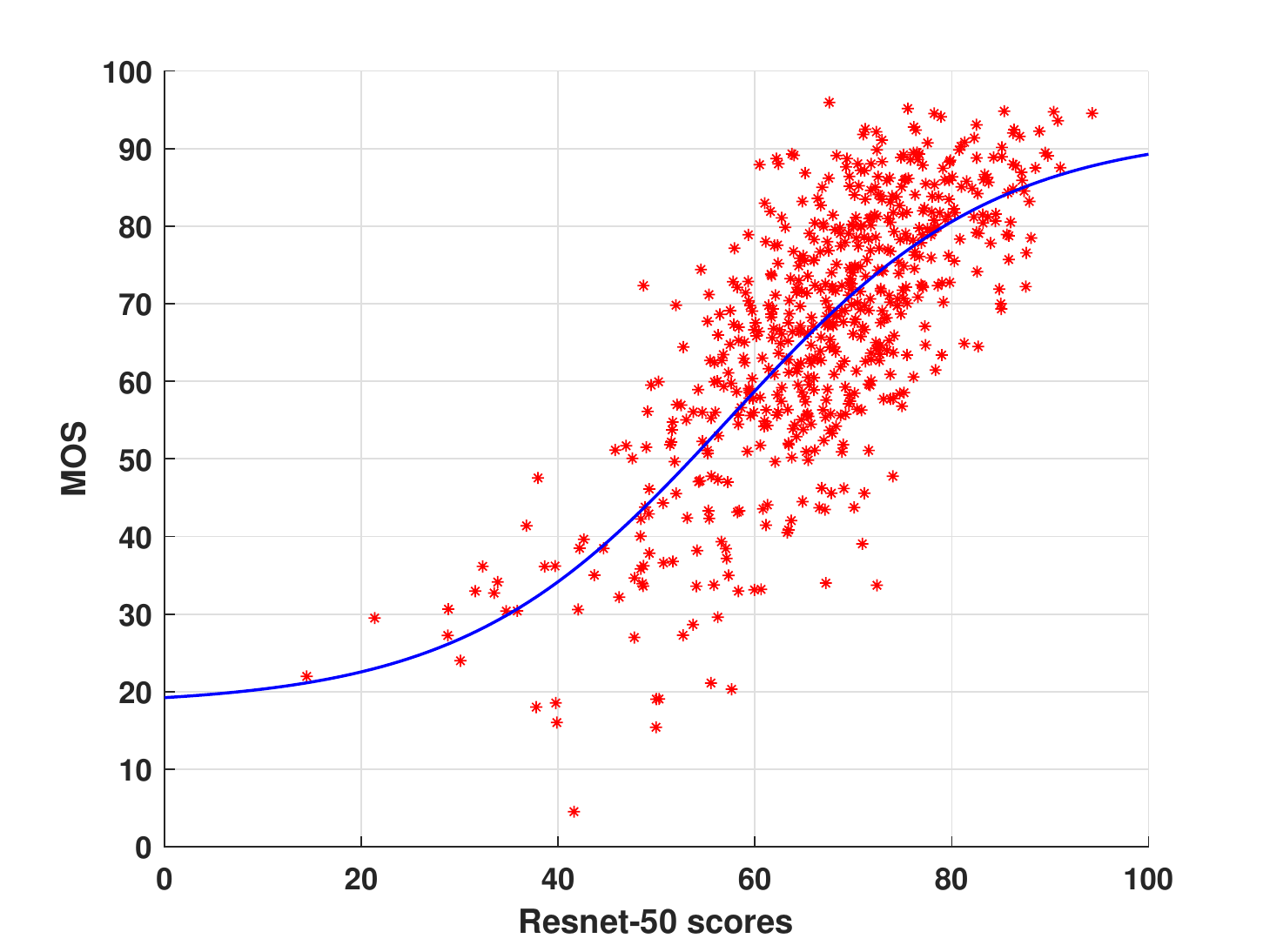}}
 \subfigure[]{
   \label{performance_scatter:proposed} 
   \includegraphics[width=0.48\columnwidth]{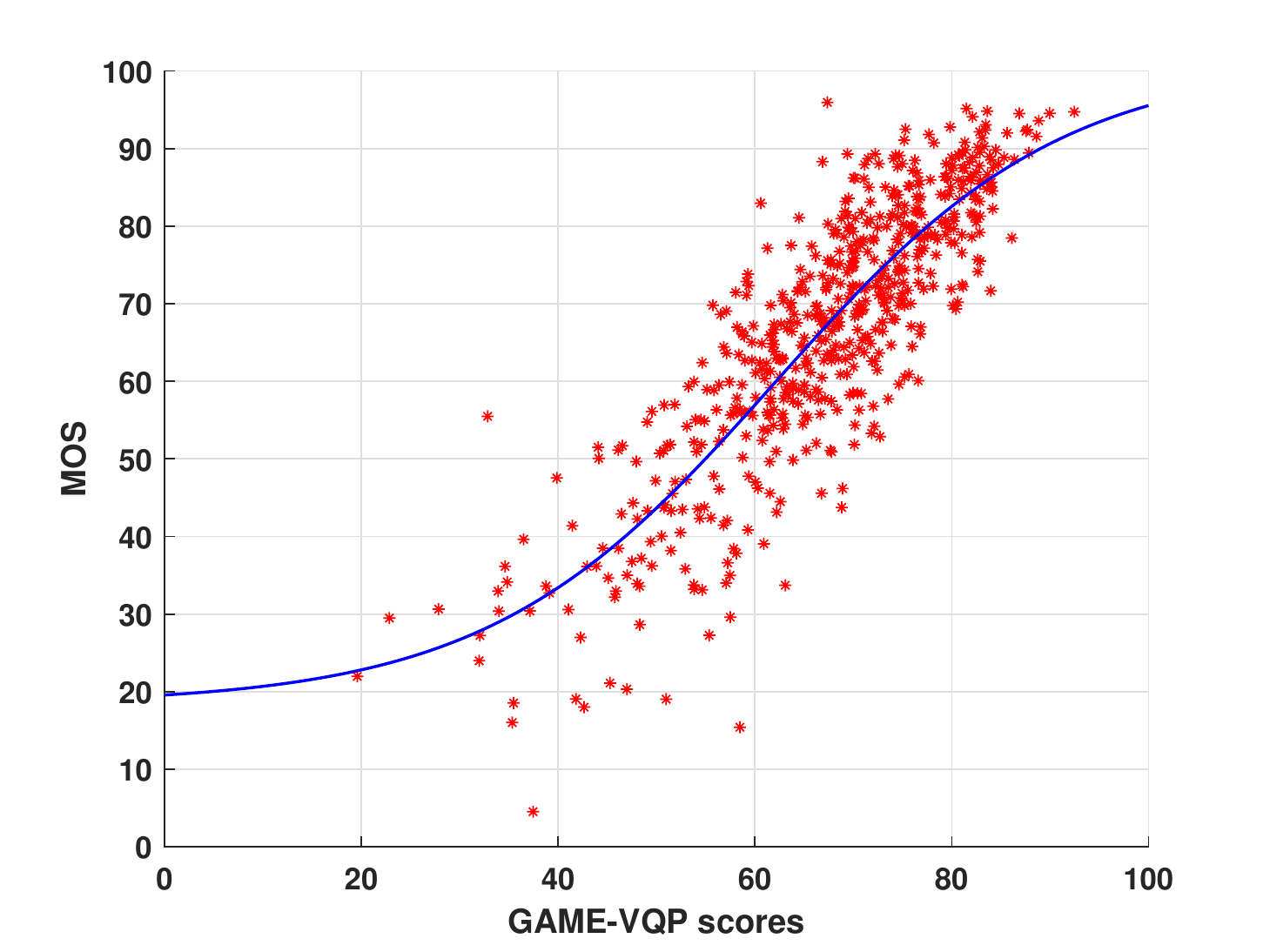}}   
 \caption{Scatter plots of predicted quality scores versus MOS trained with an SVR using 5-fold cross validation on all videos in the LIVE-YouTube Gaming Video Quality Database. (a) NIQE, (b) TLVQM, (c) RAPIQUE, (d) Resnet-50, (e) GAME-VQP. }
 \label{performance_scatter} 
\end{figure}

Scatter plots of VQA model predictions are a good way to visualize model correlations. 
To calculate scatter plots over the entire LIVE-YT-Gaming database, we applied 5-fold cross validation and aggregated the predicted scores obtained from each fold. 
Scatter plots of MOS versus the quality predictions produced by NIQE, TLVQM, RAPIQUE, Resnet-50 and our proposed model are given in Fig. \ref{performance_scatter}. 
As may be observed in Fig. \ref{performance_scatter:NIQE}, the predicted NIQE scores correlated poorly with the MOS. 
The other three models, as shown in Figs. \ref{performance_scatter:TLVQM}, \ref{performance_scatter:RAPIQUE} and \ref{performance_scatter:Resnet50}, followed regular trends against the MOS. 
As may be clearly seen in Fig. \ref{performance_scatter:proposed}, the distribution of the proposed model predictions are more compact than those of the other models/modules, in agreement with the higher correlations against MOS that it attains.

\section{Conclusion}
\label{conclusion}

UGC gaming videos are receiving significant consumer interest. 
We have presented a new gaming video quality assessment database, called LIVE-YT-Gaming, containing 600 videos of unique user generated gaming contents labeled by 18,600 subjective ratings. 
We designed and conducted a new online study to collect subjective data from a smaller group of reliable subjects. 
As compared against laboratory and crowdsourced studies, our online study has several advantages. 
We also tested several popular general-purpose and gaming-specific video quality assessment models on the new database, and compared their performances with that of a new VQA model, called GAME-VQP designed for gaming videos. 
We show that while some existing VQA models performed well, GAME-VQP significantly outperformed the others. 
We believe this new subjective data resource will help other researchers advance their work on the VQA problem for gaming videos.


%



\section*{Acknowledgment}

The human study was approved by the Institutional Review Board of UT-Austin. 
This research was supported by a gift from YouTube, and by grant number 2019844 for the National Science Foundation AI Institute for Foundations of Machine Learning (IFML). 

\ifCLASSOPTIONcaptionsoff
  \newpage
\fi



%
%
%

\bibliographystyle{IEEEtran}
\bibliography{0_main}{}

%

\end{document}